%% file: main.tex
\documentclass{article}

\usepackage[utf8]{inputenc}
\usepackage[T1]{fontenc}

\usepackage{microtype}
\usepackage{graphicx}
\usepackage{subcaption}
\usepackage{booktabs} % for professional tables

\usepackage{hyperref}

\usepackage[accepted]{icml2026}

\usepackage{amsmath}
\usepackage{amssymb}
\usepackage{mathtools}
\usepackage{amsthm}

\usepackage{cuted}
\usepackage{capt-of}

\usepackage[table,dvipsnames]{xcolor}
\usepackage{url}
\usepackage{wrapfig}
\usepackage{enumitem}
\usepackage{makecell}
\usepackage{multirow}
\usepackage[capitalize,noabbrev]{cleveref}

\usepackage{fontawesome5}   % gives you \faGithub

\newcommand{\eg}{\emph{e.g}\onedot}

\makeatletter
\DeclareRobustCommand\onedot{\futurelet\@let@token\@onedot}
\def\@onedot{\ifx\@let@token.\else.\null\fi\xspace}
\makeatother
\usepackage{xspace}

\theoremstyle{plain}
\newtheorem{theorem}{Theorem}[section]
\newtheorem{proposition}[theorem]{Proposition}

\theoremstyle{definition}

\theoremstyle{remark}

\icmltitlerunning{CPPO: Contrastive Perception Policy Optimization for VLM Agents}

\begin{document}

% \twocolumn[
%   \icmltitle{CPPO: Contrastive Perception Policy Optimization for VLM Agents}

%   % It is OKAY to include author information, even for blind submissions: the
%   % style file will automatically remove it for you unless you've provided
%   % the [accepted] option to the icml2026 package.
%   \icmlsetsymbol{equal}{*}

%   \begin{icmlauthorlist}
%     \icmlauthor{Anonymous Authors}{anon}
%   \end{icmlauthorlist}

%   \icmlaffiliation{anon}{Anonymous Institution}

%   \icmlcorrespondingauthor{Anonymous Authors}{anon@example.com}

%   \icmlkeywords{Reinforcement Learning, Vision Language Models, Multimodal Agents,
%     Contrastive Learning, Perception, Agent Robustness}

%   \vskip 0.3in
% ]

% % This command actually creates the footnote in the first column listing the
% % affiliations and the copyright notice.

\twocolumn[
  \icmltitle{CPPO: Contrastive Perception Policy Optimization for VLM Agents}
  \icmlsetsymbol{equal}{*}
  \icmlsetsymbol{corresp}{\dag}
  \icmlsetsymbol{lead}{\ddag}
  \begin{icmlauthorlist}
    \icmlauthor{Ahmad Rezaei}{corresp,huawei}
    \icmlauthor{Mohsen Gholami}{huawei}
    \icmlauthor{Saeed Ranjbar Alvar}{huawei}
    \icmlauthor{Kevin Cannons}{huawei}
    \icmlauthor{Mohammad Asiful Hossain}{huawei}
    \icmlauthor{Zhou Weimin}{cloud}
    \icmlauthor{Yong Zhang}{lead,huawei}
    \icmlauthor{Mohammad Akbari}{lead,huawei}
  \end{icmlauthorlist}
  \icmlaffiliation{huawei}{Huawei Technologies Canada Co. Ltd.}
  \icmlaffiliation{cloud}{Huawei Cloud}
  \icmlcorrespondingauthor{Ahmad Rezaei}{ahmad.rezaei2@huawei.com}
  \icmlkeywords{Reinforcement Learning, Vision Language Models, Multimodal Agents,
    Contrastive Learning, Perception, Agent Robustness}
  
  \begin{center}
    \href{https://github.com/vbdi/cppo}{\faGithub\ Code} \quad
    \href{https://huggingface.co/collections/vbdai/cppo}{\raisebox{-0.2em}{\includegraphics[height=1em]{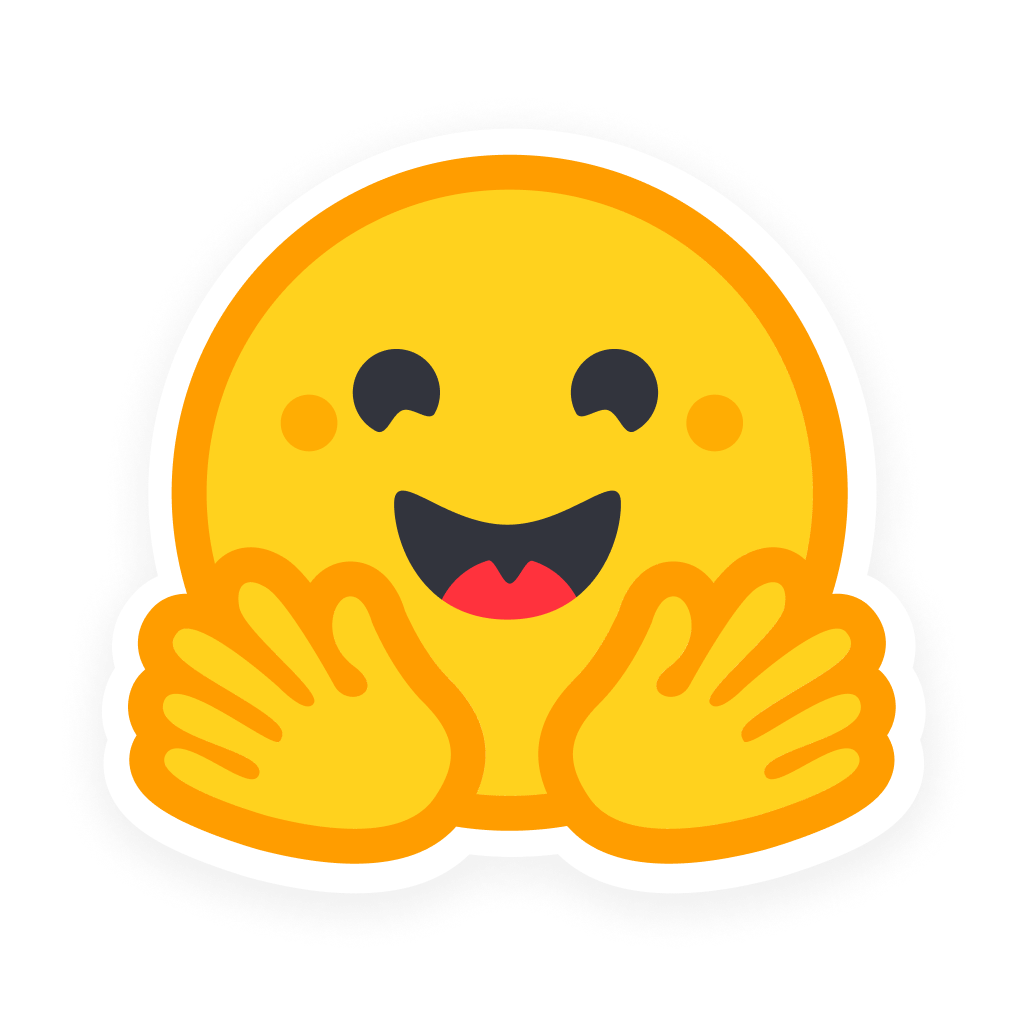}}\ CPPO}
  \end{center}
  
  \vskip 0.3in
]
% affiliations and the copyright notice.
\printAffiliationsAndNotice{\dag\ Corresponding author.\ \ddag\ Co-leads.}

\begin{figure*}
%\vspace{-50pt}
%\begin{strip}    
    \centering    
    \includegraphics[width=0.95\linewidth]{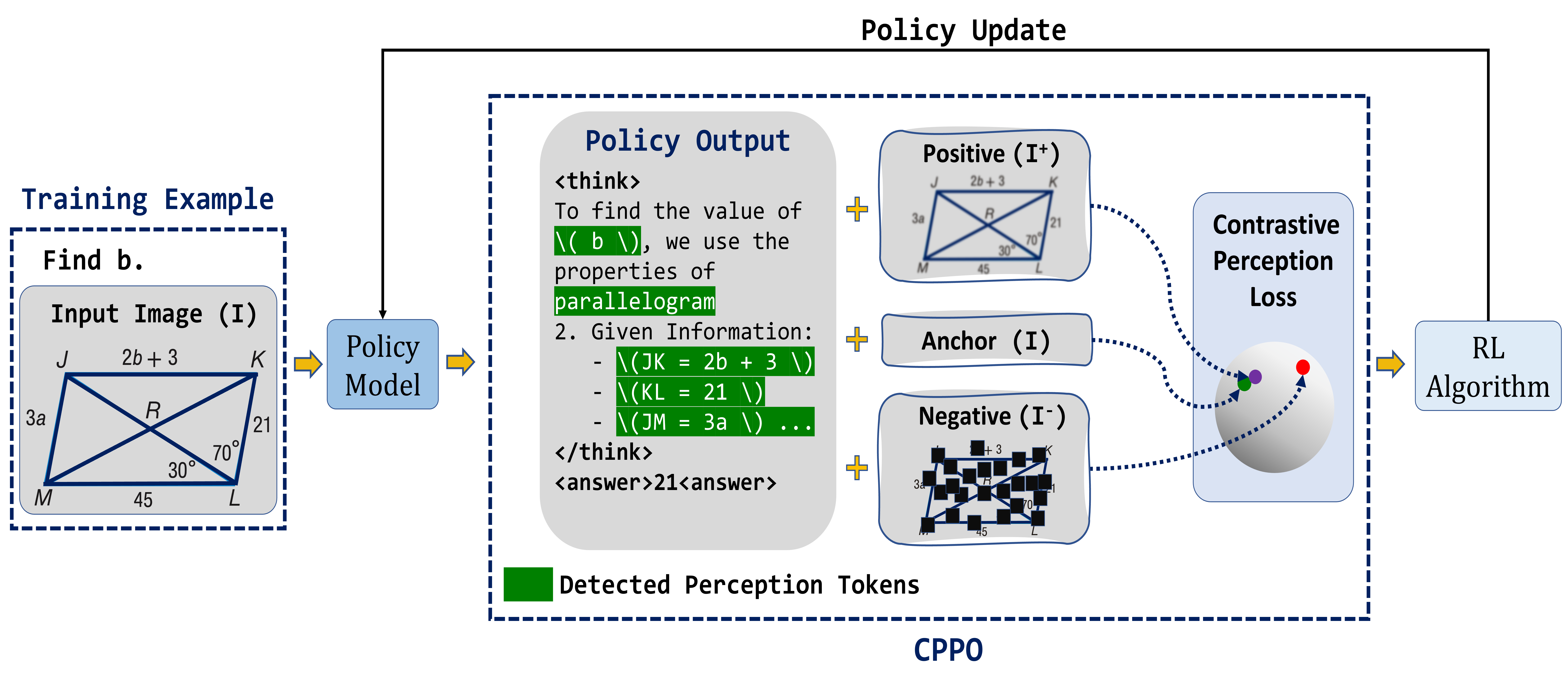}
    \captionof{figure}{Illustration of the proposed \textbf{CPPO modification to the RL objective}. CPPO augments the RL objective with a perception-specific self-supervised contrastive loss and selectively applies it to tokens that are detected as perception-dependent. Improved perception is critical for vision--language agents that must \emph{act} reliably in open-ended environments based on what they see.}
    \label{fig:teaser}
%\end{strip}
\end{figure*}

\input{0_abstract}
\input{1_intro}

\input{2_related_work}

\input{3_method}
\input{4_experiments}
\input{5_conclusion}

\newpage
\section*{Impact Statement}
This paper presents work whose goal is to advance the field of Machine Learning, with particular focus on improving the perceptual reliability of vision--language model (VLM) agents deployed in open-ended environments. Better visual grounding directly reduces a known failure mode for agents that must reason and act in open-ended environments: hallucinated or careless perception of the scene. We see no specific risks of this work beyond those already associated with the broader development of capable multimodal agents.

\bibliographystyle{icml2026}
\bibliography{main}

%%%%%%%%%%%%%%%%%%%%%%%%%%%%%%%%%%%%%%%%%%%%%%%%%%%%%%%%%%%%%%%%%%%%%%%%%%%%%%%
% APPENDIX
%%%%%%%%%%%%%%%%%%%%%%%%%%%%%%%%%%%%%%%%%%%%%%%%%%%%%%%%%%%%%%%%%%%%%%%%%%%%%%%

\clearpage
\appendix

\input{X_appendix}

\end{document}

%% file: 0_abstract.tex
\begin{abstract}
We introduce \textbf{CPPO}, a Contrastive Perception Policy Optimization method for finetuning vision--language models (VLMs). Reliable perception is a core requirement for VLM-based agents that must reason and act in open-ended environments: faulty visual grounding cascades directly into faulty actions, hallucinated tool calls, and unsafe decisions. While reinforcement learning (RL) has significantly improved reasoning in language models, extending these advances to multimodal agents requires improving both perception and reasoning. Prior works address this challenge mainly through explicit perception rewards, which often require extra LLM judges, ground-truth annotations, or forced separation of perception from reasoning. CPPO addresses this limitation in a self-supervised manner by extending the RL objective with a Contrastive Perception Loss (CPL) that provides a direct learning signal for visual grounding. The contrastive objective encourages the model to become more sensitive to input visual information. To apply this signal effectively, CPPO identifies perception tokens using an entropy-shift mechanism in the model's output distributions under perturbed images and applies the contrastive loss selectively to those tokens during training. Experiments show that CPPO surpasses prior methods while avoiding extra models, making training more efficient and scalable, and yielding policies that are better suited to perception-critical agentic tasks.
\end{abstract}

%% file: 1_intro.tex
\section{Introduction}
Vision--language models (VLMs) are increasingly deployed as the perceptual backbone of \emph{multimodal agents} that browse interfaces, operate computer-use environments, and ground long-horizon plans in real visual scenes. In all such settings, the agent's behavior is only as trustworthy as its perception: a single misread chart, a misidentified UI element, or a hallucinated visual fact propagates into wrong actions and unsafe outcomes. Improving the perceptual grounding of VLM policies is therefore a core safety lever for agents in open-ended environments.

Reinforcement learning (RL) with verifiable rewards has emerged as an effective finetuning method. Notably, \cite{deepseekr1} showed the potential of language models to develop reasoning capabilities without explicit step-by-step supervision, focusing on their self-evolution through a pure RL process. In contrast, VLMs often exhibit weaker multimodal reasoning performance compared to their language-only counterparts \citep{yang2025r1,bi2025reasoning,gholami2025spatial,cannons2025segments}. Given the success of RL in language models, recent research has focused on extending RL-based methods to VLMs and multimodal reasoning \citep{visionary-r1,PAPO,vision_matters,noisyrollout}.

% Reinforcement learning (RL) with verifiable rewards has emerged as an effective finetuning method for language and multimodal policies. Notably, \citet{deepseekr1} showed the potential of language models to develop reasoning capabilities without explicit step-by-step supervision, focusing on their self-evolution through a pure RL process.
% Given the success of RL in language models, recent research has focused on extending this approach to vision-language models (VLMs) and multimodal reasoning \cite{visionary-r1,PAPO,vision_matters,noisyrollout}.

In the language-only setting, the policy model draws on its internal knowledge to generate step-by-step logical inference tokens, which we refer to as reasoning tokens. For a VLM policy, however, accurate perception is also required to generate query-relevant factual tokens from the image. We refer to these tokens that encode image information as perception tokens.
PAPO \cite{PAPO} shows that wrong perception tokens are a significant source of failures in multimodal reasoning.
However, RL algorithms with verifiable final-answer rewards (\eg, \citet{deepseekr1}) do not separate perception from reasoning errors.
This design is problematic, since inaccurate perception tokens will lead to an incorrect final answer, even with correct reasoning steps.
Therefore, achieving the optimal policy is difficult when all output tokens are penalized based on the final answer alone.
This limitation raises two questions: \textit{1) How can the output perception and reasoning tokens be disentangled for a VLM policy? 2) How to best define an explicit perception loss/reward?}

To address the first question, \citet{visionary-r1} and \citet{ssr1} force the policy to separate perception from reasoning by defining specific generation tags: \texttt{<perception>} and \texttt{<think>}. However, forcing such a separation disrupts the natural reasoning process of the model, making it difficult to apply to many tasks (\eg, with complex images). In addition, the process becomes vulnerable to reward hacking (where the model places the final answer in the perception section to maximize reward).
Thus, we argue that perception and reasoning should be disentangled within the model's natural generation flow.

In order to address the second question, Visionary-R1 \cite{visionary-r1}, Vision-SR1 \cite{ssr1}, and Perception-R1 \cite{perception-r1} rely on an LLM and utilize either the policy's own perceptual outputs or ground-truth Chain-of-Thought (CoT) annotations to compute perception rewards.
Such evaluation of perception outputs with LLMs still requires explicit separation of perception from reasoning, incurs computational overhead, and relies on unscalable CoT supervision.
PAPO \cite{PAPO} takes a different approach via a KL divergence loss between model outputs conditioned on the original and corrupted versions of the images. However, the KL divergence is unbounded, which can easily cause reward collapse and makes the method's hyperparameters sensitive. Moreover, PAPO applies the perception loss uniformly across all tokens and output rollouts, regardless of whether they correspond to perception or reasoning, or whether the outputs are correct or incorrect. Applying divergence over reasoning tokens leads to over-regularization, while maximizing divergence on wrong perception tokens effectively reinforces incorrect perception outputs.

Motivated by these observations, we propose \textbf{C}ontrastive \textbf{P}erception \textbf{P}olicy \textbf{O}ptimization (CPPO), an RL solution designed for VLMs.
CPPO integrates two main components into the training process: (1) a mechanism that uses the policy's own output probability distribution to determine the tokens in a generated response that the policy most strongly considers as perception tokens in its current state, and (2) a token-level Contrastive Perception Loss (CPL) incorporated into the RL objective to enforce \emph{differential sensitivity} to vision information. Specifically, in each training step, we compare the policy's entropy for each token within responses when the policy is conditioned on the original image as well as a perturbed image with information-removing augmentations. Tokens whose entropy increases the most under this perturbation are selected as perception tokens by the policy, since their distribution exhibits the highest mutual information with the image.

After identifying vision-dependent tokens in the policy's output, we compute the token-level CPL term. Unlike prior work, CPL is an \textit{unsupervised} perception contrastive loss that does not require additional CoT supervision or proprietary models. Specifically, for each input image, we create two other variants: an information-preserving perturbation that retains query-relevant content and an information-removing perturbation that obscures such information. CPL is then implemented as an InfoNCE contrastive loss \cite{chen2020simple}: the token probability distribution conditioned on the original image serves as the anchor, the distribution under the information-preserving perturbation as the positive, and the distribution under the information-removing perturbation as the negative sample.
Crucially, the contrastive loss is applied only to perception tokens from \emph{correct} rollouts, ensuring that anchors correspond to accurate and verified perception tokens. This provides targeted perception feedback to the policy, thereby improving its visual grounding capability.

\textbf{Relevance to agents in the wild.} Although our experiments target standard multimodal reasoning benchmarks, the failure mode that CPPO targets---ungrounded or weakly-grounded perception tokens---is precisely the failure mode that compromises VLM agents acting on real visual inputs. A policy that has been explicitly optimized to be \emph{differentially sensitive} to visual content is, by construction, less likely to invent visual facts when its observations are noisy, partially occluded, or adversarially perturbed---all of which are routine conditions in deployed agentic settings.

In summary, the major contributions of our work are as follows:
\begin{itemize}[leftmargin=*]
    \item We propose CPPO, an RL-based finetuning solution tailored for VLMs to disentangle perception and reasoning improvement of the policy, motivated by the perceptual reliability requirements of VLM agents.
    \item We propose CPL, an \textit{unsupervised} perception-specific contrastive loss to optimize a VLM policy.
    \item We propose an entropy-based perception token detection method, where the VLM policy identifies its own perception tokens using its output distribution.
    \item We show the superiority of CPPO compared with prior perception-specific RL methods across math and visual reasoning benchmarks.
\end{itemize}

%% file: 2_related_work.tex
\section{Related Work}

% \textbf{RL for Post-training.}
In this section, we categorize the related RL methods proposed for VLMs into three directions: 1) sampling and rollout augmented methods, 2) RL combined with SFT or off-policy data, and 3) perception-aware approaches. Our approach falls into the third category, while the other directions are orthogonal to our method. We also discuss the background of using contrastive learning in RL.
%We discuss three \kevin{directions that have been considered for using RL with VLMs}: \saeed{@kevin: Third group is not for VLMs} 1) sampling and rollout augmented methods 2) RL combined with SFT or off-policy data 3) Perception-aware methods. 

\textbf{Sampling and Rollout Augmented RL with VLMs.} This line of work improves robustness and training efficiency by mixing trajectories from clean and moderately distorted images during RL training. NoisyRollout \cite{noisyrollout} and Vision Matters \cite{vision_matters} use input perturbations to stabilize grounding and enhance generalization. Shuffle-R1 \cite{shuffler1} introduces pairwise trajectory sampling and advantage-based batch reshuffling to improve gradient signal quality and increase exposure to valuable rollouts. VL-Rethinker \cite{vl-rethinker} proposes selective sample replay to address the “vanishing advantages” problem and forced rethinking, which appends a trigger token to enforce self-reflective reasoning. This line of work is orthogonal to CPPO.

\textbf{RL Combined with SFT or Off-Policy with VLMs}. This line of research combines on-policy RL with off-policy CoT or SFT training. Vision-R1 \cite{vision-r1}, Look-back \cite{lookback}, OpenVLThinker \cite{openvlthinker}, VisionThink \cite{visionthink}, and \citet{semi-off-policy} focus on semi-off-Policy RL with emphasis on rethinking, iterative pipelines, or off-policy data to enhance slow-thinking reasoning and overall training stability. Similar to the prior category, this line of work is also orthogonal to our work.

\textbf{Perception-Aware RL with VLMs.}
This line of work aims to improve how VLM policies couple visual perception with reasoning. One direction adopts decoupled architectures, using a VLM for visual description and an LLM for reasoning, optimized jointly with RL \cite{guo2025decoupled,huawei2025decoupled}. Another approach explicitly separates perception from reasoning tokens in the model output. Vision-SR1 \cite{ssr1} and Visionary-R1 \cite{visionary-r1} enforce special tags (e.g., <perception> and <think>) and use an LLM to evaluate perception tokens, while Perception-R1 \cite{perception-r1} instead leverages supervised CoT trajectories to assess perception within reasoning paths. However, these methods rely on additional models or supervision, increasing computational cost and limiting scalability. To avoid extra supervision, PAPO \cite{PAPO} introduces an unsupervised KL divergence loss between outputs conditioned on original and corrupted images. However, the unbounded KL objective can cause instability and is applied uniformly across all tokens, potentially over-regularizing reasoning tokens. Recent work further studies token-level perception in multimodal RL, showing that only a subset of tokens strongly depend on visual evidence and that perception signals vary across trajectories \cite{huang2025spotlight}.

\textbf{Contrastive Learning in RL.}
Contrastive learning has been explored in general RL literature to improve representation quality and sample efficiency, e.g., CURL \cite{curl}, SPR \cite{spr}, SODA \cite{soda}, and TACO \cite{taco}.   
Recently, contrastive methods have been adopted for preference alignment in LLMs, e.g., Contrastive Preference Learning \cite{conpref} and Contrastive Preference Optimization (CPO) \cite{cpo}. However, VLM policy optimization remains unexplored. Our approach introduces a token-level contrastive loss as a self-supervised approach to improve policy's perception quality.
% alignment with human feedback. Contrastive Preference Learning \cite{conpref} proposes learning directly from human feedback signals without relying on standard RLHF pipelines, by using a contrastive objective to distinguish preferred behaviors. Similarly, Contrastive Preference Optimization (CPO) \cite{cpo} applies this principle in the context of LLMs, showing that contrastive objectives can outperform traditional RL-based preference optimization in domains like machine translation. While these methods highlight the versatility of contrastive learning across RL and alignment, VLM policy optimization remains unexplored. Our approach introduces a \emph{token-level contrastive loss} tailored to VLMs, that is applied specifically to vision-dependent tokens within reasoning rollouts. 

\textbf{Augmenting Input Image.} Adding perturbation to image inputs is used in prior works to reduce object-hallucination (VCD by \citet{leng2024mitigating}) and improve visual grounding (SeVa by \citet{zhu2024self}) at inference/alignment time. Other approaches improve vision encoder of VLMs (Epistemic-Uncertainty Masking by \citet{seo2025epistemic}) or inject image-grounded signals at decode time (MARINE by \citet{zhao2024mitigating}). Unlike these methods, which are primarily training-free, post-hoc, or operate at the vision encoder level, CPPO embeds image perturbations into RL training and applies a token-level contrastive loss only to entropy-identified perception tokens to shape the policy’s visual sensitivity during optimization rather than correcting it post hoc.

%% file: 3_method.tex
\section{Method}
In this section, we review RL with verifiable rewards and then elaborate our proposed unsupervised contrastive perception policy optimization and how it is selectively applied to perception-dependent tokens.
\subsection{Preliminaries}
\textbf{Group Relative Policy Optimization (GRPO).} GRPO~\cite{deepseekr1} includes RL fine-tuning of the policy VLM $\pi_\theta$ with parameters $\theta$ on verifiable tasks. 
%by comparing sampled trajectories within a sampled answer group. 
Given an input set $x=\{q,I\}$ including query $q$ and image $I$, a group of $G$ output trajectories (responses) $\{\mathbf{o}_1,\ldots,\mathbf{o}_G\} \sim \pi_\theta(\cdot \mid x)$ are sampled. Each output $\mathbf{o}_i$ consists of $T$ tokens $\{o_{i,1},\ldots,o_{i,t},\ldots,o_{i,T}\}$ and receives a scalar reward $R(\mathbf{o}_i)$, typically reflecting correctness.
Relative advantages are computed as:
\begin{equation}
A_{i} = \frac{R(\mathbf{o}_i) - mean\big(R(\mathbf{o}^{1:G})\big)}{std\big(R(\mathbf{o}^{1:G})\big)},
\label{eq:relative_advantage}
\end{equation}
where $i \in [1,G]$. The GRPO objective is then defined as:
\begin{equation}
\small
\begin{aligned}
\mathcal{J}_{\mathrm{GRPO}}(\theta) &= \mathbb{E}_{\mathbf{o}_i \sim \pi_{\theta_{old}}} \tfrac{1}{G} \sum_{i=1}^G \tfrac{1}{|\mathbf{o}_i|} \sum_{t=1}^{|\mathbf{o}_i|} \Big\{\\
&\quad \min\!\big(r_{i,t}(\theta) A_i,\, \mathrm{clip}(r_{i,t}(\theta), 1-\epsilon, 1+\epsilon)\, A_i\big) \\
&\quad - \beta\, \mathrm{KL}[\pi_\theta \,\|\, \pi_{ref}]\Big\},
\end{aligned}
\label{eq:grpo_loss}
\end{equation}
where $r_{i,t}(\theta) = \pi_\theta(o_{i,t}\mid x,\mathbf{o}_{i,<t}) / \pi_{\theta_{old}}(o_{i,t}\mid x,\mathbf{o}_{i,<t})$ is the importance ratio. The KL penalty controls the deviation from the frozen reference policy $\pi_{ref}$ with weight $\beta$. Output trajectories are generated by the rollout policy $\pi_{\theta_{old}}$, and the hyperparameter $\epsilon$ controls clipping large policy updates. %All tokens generated contribute equally to this term. While GRPO improves reasoning ability, its reward signal only evaluates the final output correctness and does not distinguish between reasoning error and vision mistakes.
In this setting, the correctness reward alone provides no explicit signal to enhance the policy model’s perceptual sensitivity. Our CPL loss aims to address this gap.

%%%%%%%%%%%%%%%%%%%%%%%%%%%%%%%%%%%%%%%%%%%%%%%%%%%%%%%%%%%%%%%%%%
\subsection{CPPO: Contrastive Perception Policy Optimization}
\begin{figure*}[t!]
    \centering
    \includegraphics[width=0.9\linewidth]{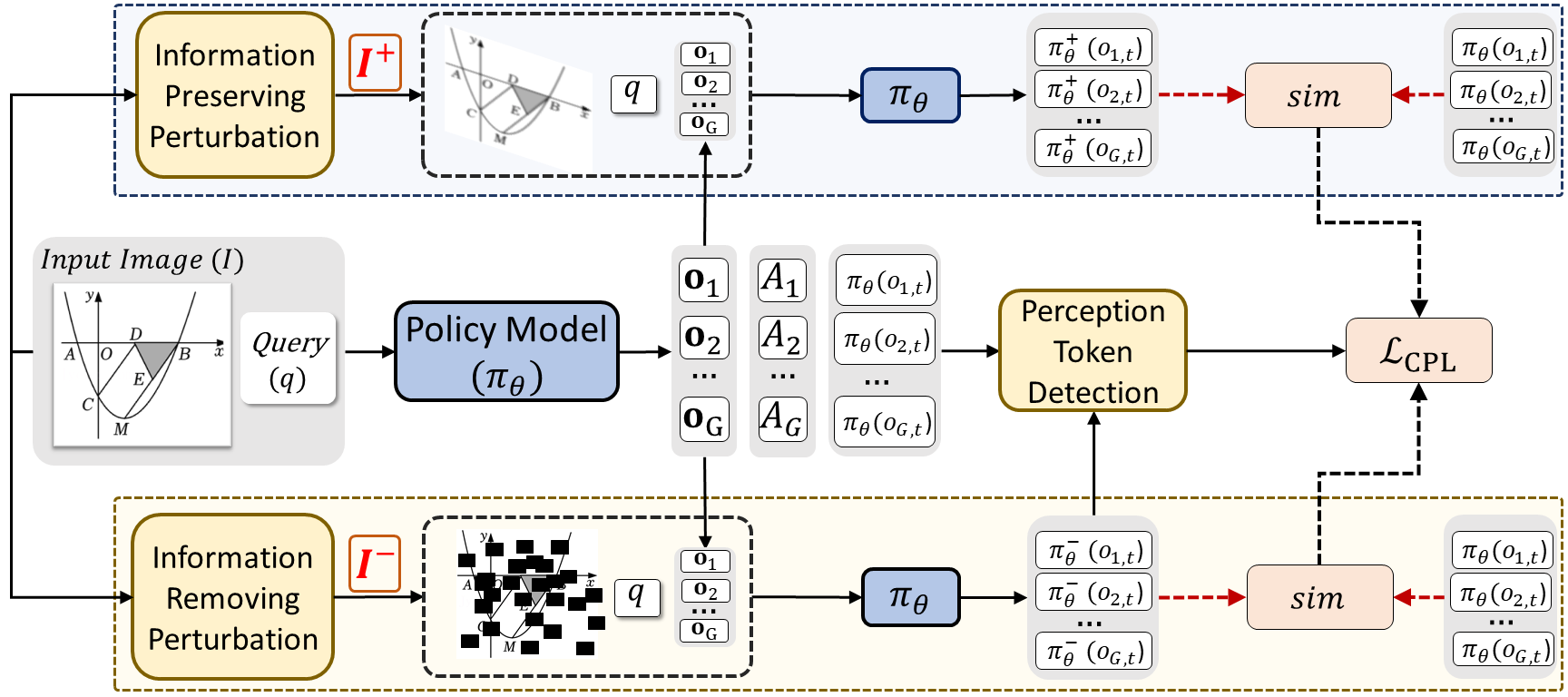}
    \caption{ 
    \textbf{Detailed overview of CPPO.} For each rollout $\mathbf{o}_i$, perception tokens are identified and their probability distributions are computed under three conditions: the original image $I$ (anchor sample: $\pi_{\theta}(o_{i,t})$), an information-preserving perturbation $I^{+}$ (positive sample: $\pi_{\theta}^+(o_{i,t})$), and an information-removing perturbation $I^{-}$ (negative sample: $\pi_{\theta}^-(o_{i,t})$). Similarities $sim\big(\pi_{\theta}(o_{i,t}), \pi_{\theta}^+(o_{i,t})\big)$ and $sim\big(\pi_{\theta}(o_{i,t}), \pi_{\theta}^-(o_{i,t})\big)$ are computed and incorporated into the Contrastive Perception Loss (CPL), which serves as an additional perception-specific term in the RL objective. Notations are simplified for brevity.
    }
    \label{fig:method}
\end{figure*}
While prior work has explored guiding the policy model toward improved perceptual understanding by providing explicit vision rewards, our approach instead augments the RL objective function with a perception-dependent contrastive loss. \Cref{fig:method} and \cref{alg:cppo} illustrate the overall framework. Inspired by contrastive representation learning \cite{chen2020simple}, the central idea of CPPO is to encourage the policy to be \emph{differentially sensitive} to visual perturbations in the input image at the \emph{token level} by selectively applying a contrastive perception loss to perception-dependent tokens.

\begin{algorithm}[t]
{\small
\caption{CPPO}%: Contrastive Perception Policy Optimization}
\label{alg:cppo}
\begin{algorithmic}[1]

\REQUIRE policy $\pi_\theta$, rollout policy $\pi_{\theta_{\rm old}}$, dataset $\mathcal{D}$, RL objective $\mathcal{J}$, contrastive perception loss weight $\lambda$, top-$k$ ratio $k$

\FOR{each training step}
    \STATE Sample input $(q, I) \sim \mathcal{D}$
    \STATE Generate rollout $o_i \sim \pi_{\theta_{\rm old}}(\cdot \mid q,I)$
    \STATE Compute reward $R(o_i)$ and advantage $A_i$
    % \Statex
    % \State 
    \FOR{each rollout $o_i$} %\Comment{Perception Token Detection}
        \STATE Create information-removing image $I^{-}$
        \STATE Compute entropy increase $\Delta H_{i,t}$ for all tokens
        \STATE Select top-$k$ positive $\Delta H_{i,t}$ tokens: $\mathcal{S}_{\rm perception}(o_i)$
    \ENDFOR
    % \Statex
    % \State 
    \FOR{each rollout $o_i$} %\Comment{Contrastive Perception Loss}
        \STATE Create information-preserving image $I^{+}$
        \STATE For all $t \in \mathcal{S}_{\rm perception}(o_i)$:
        \STATE \qquad Compute contrastive loss $\mathcal{L}^{\rm InfoNCE}_{i,t}$
        \STATE $\mathcal{L}_{\rm CPL}(o_i) = \frac{1}{|\mathcal{S}_{\rm perception}(o_i)|}\sum_{t}\mathcal{L}^{\rm InfoNCE}_{i,t}$
    \ENDFOR
    \STATE $\mathcal{J}(\theta) = \mathcal{J}(\theta) 
            - \lambda \cdot \mathbf{1}[A_i>0]\, \mathcal{L}_{\rm CPL}(o_i)$ %\Comment{Combine With RL Objective With Advantage Gating}
    \STATE Update policy $\theta \gets \theta +  \nabla_\theta \mathcal{J}(\theta)$
    \STATE Update rollout policy $\theta_{\rm old} \gets \theta$
\ENDFOR
\end{algorithmic}} 
\end{algorithm}

% \begin{algorithm}[t]
% {\small
% \caption{CPPO}%: Contrastive Perception Policy Optimization}
% \label{alg:cppo}
% \begin{algorithmic}[1]
% \REQUIRE policy $\pi_\theta$, rollout policy $\pi_{\theta_{\rm old}}$, dataset $\mathcal{D}$, RL objective $\mathcal{J}$, CPL weight $\lambda$, top-$k$ ratio $k$
% \FOR{each training step}
%     \STATE Sample input $(q, I) \sim \mathcal{D}$
%     \STATE Generate rollouts $\{o_i\} \sim \pi_{\theta_{\rm old}}(\cdot \mid q,I)$ and compute rewards $R(o_i)$ and advantages $A_i$
%     \STATE \textit{// Perception Token Detection}
%     \STATE Create information-removing image $I^{-}$
%     \STATE Compute entropy increase $\Delta H_{i,t}$ for all $i,t$
%     \STATE Select top-$k$ positive $\Delta H_{i,t}$ tokens: $\mathcal{S}_{\rm perception}(o_i)$ for each $i$
%     \STATE \textit{// Contrastive Perception Loss}
%     \STATE Create information-preserving image $I^{+}$
%     \STATE For each $i$ and each $t \in \mathcal{S}_{\rm perception}(o_i)$, compute $\mathcal{L}^{\rm InfoNCE}_{i,t}$
%     \STATE $\mathcal{L}_{\rm CPL}(o_i) = \frac{1}{|\mathcal{S}_{\rm perception}(o_i)|}\sum_{t}\mathcal{L}^{\rm InfoNCE}_{i,t}$
%     \STATE \textit{// Combine with RL objective + advantage gating}
%     \STATE $\mathcal{J}(\theta) \mathrel{+}= -\lambda \cdot \frac{1}{G}\sum_i \mathbf{1}[A_i>0]\, \mathcal{L}_{\rm CPL}(o_i)$
%     \STATE Update policy $\theta \gets \theta +  \nabla_\theta \mathcal{J}(\theta)$; rollout policy $\theta_{\rm old} \gets \theta$
% \ENDFOR
% \end{algorithmic}}
% \end{algorithm}

\textbf{Contrastive Perception Loss (CPL).}
Given the generated response {$\mathbf{o}_i$} for the input $x=\{q, I\}$, CPL operates by contrasting the policy’s probability distribution of each token under perturbed views of the input image $I$. For each token $o_{i,t}$, CPL recomputes the probability distributions under two variants of the anchor image $I$:
\begin{itemize}[leftmargin=*]
    \item \textit{Information-removing perturbations} $I^{-}$, obtained from transformations such as region masking or deletion of critical visual elements that obscure query-relevant information. The policy’s output distribution under 
    $I^{-}$ should diverge from that of $I$.
    % $\pi_\theta(o_{i,t} \mid q,I^{-},\mathbf{o}_{i,<t})$,
    % $\pi_\theta(o_{i,t} \mid q,I,\mathbf{o}_{i,<t})$.  
    \item \textit{Information-preserving perturbations} $I^{+}$, obtained from transformations such as mild Gaussian noise or small brightness shifts that do not remove query-relevant content. The output distribution under $I^{+}$ should remain consistent with that of $I$.
    % , $\pi_\theta(o_{i,t} \mid q,I^{+},\mathbf{o}_{i,<t})$,
    % $\pi_\theta(o_{i,t} \mid q,I,\mathbf{o}_{i,<t})$. 
\end{itemize}
% CPL constrastive signal enforces stability under irrelevant perturbations and sensitivity under information-removing perturbations, in a fully unsupervised manner without relying on CoT annotations.
Formally, for a token $o_{i,t}$ in $\mathbf{o}_i$, we denote the policy probability distribution under the original image as the anchor
\begin{equation}
    \pi_\theta(o_{i,t}) = \pi_\theta(o_{i,t} \mid q,I,\mathbf{o}_{i,<t}),    
\end{equation}
the distribution under $I^{+}$ as the positive sample
\begin{equation}
\pi_\theta^{+}(o_{i,t}) = \pi_\theta(o_{i,t} \mid q,I^{+},\mathbf{o}_{i,<t}),
\end{equation}
and the distribution under $I^{-}$ as the negative sample
\begin{equation}
\pi_{\theta}^{-}(o_{i,t}) = \pi_\theta(o_{i,t} \mid q,I^{-},\mathbf{o}_{i,<t}).
\end{equation}
Let $sim(p,p^*) = -\mathrm{KL}(p \,\|\, p^*)$ denote the negative KL divergence as a similarity measure between token probability distributions, and let $s^{\pm}_{i,t} = sim\big(\pi_\theta(o_{i,t}), \pi_\theta^{\pm}(o_{i,t})\big)$. We adopt the InfoNCE loss \cite{chen2020simple} to define the contrastive objective:
\begin{equation}
\small
\mathcal{L}^{\mathrm{InfoNCE}}_{o_{i,t}}
=
- \log \frac{\exp(s^{+}_{i,t} / \tau)}{\exp(s^{+}_{i,t}/\tau) + \exp(s^{-}_{i,t}/\tau)},
\label{eq:infonce}
\end{equation}
where $\tau > 0$ is a temperature hyperparameter. Minimizing this loss encourages the anchor distribution to remain close to the positive view while being pushed away from the negative view, thereby explicitly enforcing perceptual improvement in a fully unsupervised manner without relying on CoT annotations.

\textbf{Perception Token Detection.}
Not all tokens in an output are equally dependent on perceptual input. For example, interpreting ``the base is 10 cm'' relies on visual information, whereas solving ``$x^2 + 2x + 1 = 0$'' or recalling that ``the angles of a triangle sum to $180^{\circ}$'' can be performed independently of the image. Applying CPL uniformly across all tokens may lead to excessive regularization and destabilize training. Therefore, we introduce a mechanism to selectively identify perception-dependent tokens using the model’s own output distribution, and apply CPL only to these tokens.

\begin{proposition}[\textbf{Entropy increase as a proxy for perception dependence}]
(Proof in Appendix \ref{appendix:proof-vision-entropy}) Let $I$ denote the original image, $I^{-}$ a perturbed variant that removes query-relevant perceptual information, and $\mathbf{o}_i$ a sequence of tokens generated by the policy when conditioned on $I$. The increase in entropy of a token $o_{i,t} \in \mathbf{o}_i$, when the policy is conditioned on $I^{-}$ rather than $I$, serves as a proxy for the degree to which the policy associates $o_{i,t}$ with the query-relevant visual content of $I$. The increase is calculated as:
\begin{equation}
    \Delta H_{i,t} = H(o_{i,t}|q,I^{-},\mathbf{o}_{i,<t}) - H(o_{i,t}|q,I,\mathbf{o}_{i,<t}).
\end{equation}
\label{prop: prop_1}
\end{proposition}
For token $o_{i,t}$, the predictive entropy is defined as:
\begin{equation}
\small
H(o_{i,t}|x,\mathbf{o}_{i,<t}) = - \!\!\!\sum_{o_{i,t} \in \mathcal{V}}\!\! \pi_\theta(o_{i,t} | x,\mathbf{o}_{i,<t}) \log \pi_\theta(o_{i,t} | x,\mathbf{o}_{i,<t}),
\end{equation}
where $\mathcal{V}$ denotes the vocabulary.

\begin{table*}[t!]
    \centering
    {\small
    \caption{CPPO vs. GRPO. All results are based on {avg@8}. \textbf{Bold}: the best value in each column. $\Delta_{rel}^{\%}$ shows the relative improvement over the GRPO baseline averaged over all benchmarks.}
    \label{tab:main_result_GRPO}
    \resizebox{0.95\textwidth}{!}{
    \begin{tabular}{l|ccccc|cc|cc}
         \toprule
         &\multicolumn{5}{c|}{\textbf{Math Benchmarks}}&\multicolumn{2}{c|}{\textbf{Visual Reasoning}}&\\ \cmidrule{2-10}
         \rowcolor[gray]{.9}
        \textbf{Methods} & \textbf{MVista}\textsubscript{m} & \textbf{DMath} & \textbf{WeMath} & \textbf{MVision\textsubscript{m}} & \textbf{MVerse} & \textbf{MMMU-P\textsubscript{v}}&\textbf{LogicVista}&\textbf{AVG} & $\Delta_{rel}^{\%}$  \\ \midrule
        \textit{Qwen2.5-VL-3B}& 56.4&33.7&14.5&19.5&25.7&19.9 & 32.4&28.8& -\\
        \hspace{2mm} GRPO-3B & 63.7&{45.7}&{28.4}&{25.1}&{38.3}&25.8& 37.7&37.8& -\\ \cmidrule{2-10}
         \hspace{2mm}  \textbf{CPPO-3B} & \textbf{66.3}&\textbf{48.9}&\textbf{30.8}&\textbf{25.3}&\textbf{39.4}&\textbf{28.5} &\textbf{40.9}& \textbf{40.0} & \textbf{6.0}\\ \midrule
        \textit{Qwen2.5-VL-7B} &65.6&53.2&33.3&24.5&41.2&33.7&45.1&42.3& - \\
         \hspace{2mm}  GRPO-7B&71.2&55.6&42.4&27.6& 45.0&37.9 & 47.4&{46.7} & -\\ \cmidrule{2-10}
         \hspace{2mm}  \textbf{CPPO-7B} &\textbf{72.2}&\textbf{56.9}&\textbf{44.8}&\textbf{29.9}&\textbf{46.5}&\textbf{39.0}&\textbf{48.2} & \textbf{{48.2}} & \textbf{3.7} \\
        \bottomrule
    \end{tabular}}}
\end{table*}

\textit{{Perception}-Top$k$.}
After generating $\mathbf{o}_i$ for image $I$, we construct $I^{-}$ by applying a random information-removing perturbation and compute $\Delta H_{i,t}$ for each token. Tokens are ranked by $\Delta H_{i,t}$, and the top$k$ most perception-dependent tokens are retained:
\begin{equation}
    \mathcal{S}_{\text{perception}} \;=\; \left\{\, {t} \;\middle|\; \operatorname{Rank}(\Delta H_{i,t}) \le k \cdot T \,\right\},
\label{eq:topk}
\end{equation}
where $k$ denotes the proportion of tokens with the highest entropy increase. We construct a binary mask $M_i \in \{0,1\}^T$:
\begin{equation}
    M_{i,t} =
    \begin{cases}
        1, & \text{if } t \in \mathcal{S}_{\text{perception}}, \\
        0, & \text{otherwise}.
    \end{cases}
\end{equation}

CPL is then selectively applied as:
\begin{equation}
    \mathcal{L}_{\mathrm{CPL},i,t} =
    \begin{cases}
        \mathcal{L}^{\mathrm{InfoNCE}}_{o_{i,t}} & \text{if }M_{i,t}=1, \\[6pt]
        0, & \text{if }M_{i,t}=0.
    \end{cases}
    \label{eq:CPL-loss}
\end{equation}
The overall CPL for trajectory $\mathbf{o}_i$ is:
\begin{equation}
    \mathcal{L}_{\mathrm{CPL}}(\mathbf{o}_i; I,I^{+},I^{-}) = \frac{1}{|\mathbf{o}_i|}
    \sum_{t=1}^{|\mathbf{o}_i|} \mathcal{L}_{\mathrm{CPL},i,t}.
\end{equation}

\textbf{Integration with RL Objective.}
Finally, we integrate CPL with the GRPO objective. For each sampled trajectory $\mathbf{o}_i$, we compute the standard GRPO update (\cref{eq:grpo_loss}) together with the CPL term. 

\textit{Advantage Gating Mechansim.} To prevent low-quality trajectories from introducing noisy gradients, we use an advantage gating mechanism: CPL is applied only when the trajectory’s advantage {$A_i$} is positive. Thus, the combined objective is:
\begin{equation}
\small
\begin{aligned}
\mathcal{J}(\theta) = \mathbb{E}_{\mathbf{o}_i \sim \pi_{\theta_{old}}} \Big[ &\mathcal{J}_{\mathrm{GRPO}}(\theta) \\
&\!\!\!- \lambda \tfrac{1}{G} \sum_{i=1}^G \mathbf{1}\{A_i {>} 0\}\, \mathcal{L}_{\mathrm{CPL}}(\mathbf{o}_i; I,I^{+},I^{-}) \Big],
\end{aligned}
\label{eq:objective_func}
\end{equation}
where $\mathbf{1}\{\cdot\}$ denotes the indicator function and $\lambda$ controls the strength of the contrastive objective. By incorporating advantage gating, CPL is imposed only on trajectories that outperform the group baseline, ensuring that CPL regularization reinforces successful perceptions while avoiding incorrect trajectories.

%% file: 4_experiments.tex
\section{Experiments}
\subsection{Experimental Setup}
\textbf{Training Dataset.} We train on ViRL39K \cite{vl-rethinker}, a dataset consisting of 38.8K multimodal question--answer pairs.
The dataset spans a broad range of domains, including grade school problems to broader STEM and social topics; reasoning with charts, diagrams, tables, documents, and spatial relationships.

\textbf{Evaluation.} Following prior works,
we use the following benchmarks for evaluation: LogicVista \cite{logicvista}, MathVista \cite{mathvista}, DynaMath \cite{dynamath}, WeMath \cite{wemath}, MathVision \cite{mathvision}, MathVerse \cite{mathverse}, and MMMU-Pro-Vision \cite{mmu-pro}. These benchmarks encompass math, general multimodal reasoning, and logical reasoning tasks. All evaluations are performed using VLMEvalKit \cite{vlmevalkit}. We report average accuracy@8 with an inference temperature of 1.0 to provide a more consistent and reliable measure of model performance across all the experiments in the paper.

\textbf{Baselines.} We use Qwen2.5-VL-3B and 7B \cite{Qwen2.5-VL} as the backbone models in all our experiments. We compare our CPPO with recent RL methods proposed for VLMs: OpenVLThinker-3B/7B, Visionary-R1-3B, PAPO-3B/7B, VL-ReThinker-7B, Vision-Matters-7B, NoisyRollout-7B, Perception-R1-7B, Vision-SR1-7B, and Look-Back-7B (semantic checkpoint). All of these prior works use Qwen2.5-VL-3B/7B as the policy model.

\begin{table*}[t!]
    \centering
    {\small
    \caption{CPPO vs. prior works for 3B and 7B models. All results are reported as {avg@8}.
    \textbf{Bold} indicates the best value in each column; \underline{underlined} indicates the second best.}
    \label{tab:main_result_prior_works}
    \resizebox{0.95\textwidth}{!}{
    \begin{tabular}{l|ccccc|cc|l}
         \toprule
         &\multicolumn{5}{c|}{\textbf{Math Benchmarks}}&\multicolumn{2}{c|}{\textbf{Visual Reasoning}}&\\ \cmidrule{2-9}
        \rowcolor[gray]{0.94}
        \textbf{Methods} & \textbf{MVista}\textsubscript{m} & \textbf{DMath} & \textbf{WeMath} & \textbf{MVision\textsubscript{m}} & \textbf{MVerse} & \textbf{MMMU-P\textsubscript{v}}&\textbf{LogicVista}&\textbf{AVG} \\ \midrule
        GPT4-o&60.0&34.5&47.4&30.6&41.2&51.9&52.8&{45.4}\\
        Gemini-2.0-Flash&73.4&42.1&45.8&41.3&54.6&51.7&52.3&{51.6} \\
        \midrule
         \hspace{2mm} OpenVLThinker-3B &60.0&35.6&26.3&22.3&36.9&25.0&37.4&{34.7} \\
         \hspace{2mm} Visionary-R1-3B &61.4&41.2&27.1&19.7&34.5&\underline{27.9}& 37.1&{35.5} \\
         \hspace{2mm}  PAPO-3B& \underline{64.8}&\underline{45.4}&\underline{28.1}&\underline{24.3}&\underline{38.3}&26.8 &\underline{39.4}& \underline{38.1}\\ \cmidrule{2-9}
         \hspace{2mm}  \textbf{CPPO-3B} & \textbf{66.3}&\textbf{48.9}&\textbf{30.8}&\textbf{25.3}&\textbf{39.4}&\textbf{28.5} &\textbf{40.9}& \textbf{{40.0}}\\ \midrule
         \hspace{2mm} OpenVLThinker-7B &70.7&43.9&38.4&27.5&40.7&35.5&45.8 &{43.9}\\
         \hspace{2mm} Vision-SR1-7B  &67.0&52.6&33.6&28.0&40.7&{38.9}&43.2&{43.9}\\
         \hspace{2mm} Look-Back-7B &69.1&52.5&39.8&25.8&41.9&34.5&46.3& {44.8}\\
          \hspace{2mm} Vision-Matters-7B &68.6&54.5&40.1&25.2&45.3&35.5& 45.1&{45.3}\\
          \hspace{2mm}  PAPO-7B &\underline{71.6}&54.7& 39.5&26.5&44.5&\underline{38.7}&45.8 & {46.8} \\
          \hspace{2mm} PerceptionR1-7B & 70.0&{55.8}&\textbf{45.4}&27.6&46.0&38.1 &45.5& {47.3} \\
          \hspace{2mm} NoisyRollout-7B&71.1&\underline{55.9}&44.4&\underline{29.4}&\underline{46.4}&38.5&\underline{47.9}&{\underline{47.7}}\\ \cmidrule{2-9}
         \hspace{2mm}  \textbf{CPPO-7B} &\textbf{72.2}&\textbf{56.9}&\underline{44.8}&\textbf{29.9}&\textbf{46.5}&\textbf{39.0}&\textbf{48.2} & \textbf{48.2}  \\
        \bottomrule
    \end{tabular}}}
\end{table*}

\textbf{Perturbation Types.} Information-removing perturbations include random 80\% patch-wise masking and random 30\% cropping (retaining only 30\% of the image) to obscure the majority of the visual content. For information-preserving perturbations, we apply lightweight color perturbations such as color jitter and Gaussian noise as well as mild geometry perturbations such as random perspective and random rotation. These perturbations modify the image appearance without eliminating critical information. At each training step, one augmentation is randomly sampled from each augmentation set. Samples of perturbations along with detailed parameter settings for all perturbations are provided in Appendix \ref{appendix:perturbations}.

\textbf{Implementation Details.} We use \textit{verl} \cite{verl} as our RL training framework. The policy models are initialized with Qwen2.5-VL-3B/7B. We train the policy model with GRPO and CPPO for 2 epochs on the ViRL39K dataset with a group size of 5 and a global batch size of 512. Both the vision encoder and LLM of the baselines were updated during training. For other RL-related hyperparameters, we use the default settings of \textit{verl}. Our training settings match those of PAPO for a controlled comparison. More details are in Appendix \ref{appendix:details}.

\subsection{Main Results}
\textbf{Comparison to Baseline GRPO.}
Applying CPPO to the Qwen2.5-VL-3B and -7B baselines yields consistent and substantial improvements on the test benchmarks, with average absolute performance gains of {11.2\%} and {5.9\%}, respectively. As reported in \cref{tab:main_result_GRPO}, CPPO achieves a higher accuracy than GRPO across all benchmarks---average relative gains of 6.0\% for the 3B model and 3.7\% for the 7B model.
Overall, these results confirm that CPPO is a more effective optimization strategy than GRPO, especially for mid-sized models, and establishes CPPO as a strong and scalable alternative for finetuning large VLMs. Qualitative results are given in Appendix \ref{appendix:qualitative}.

\textbf{Comparison to Other Methods.}
As shown in \cref{tab:main_result_prior_works}, CPPO consistently surpasses prior methods across all benchmarks for the 3B model. For the 7B model, CPPO also outperforms existing approaches on all benchmarks (except WeMath), demonstrating stronger generalization. In particular, when compared to PAPO---the most relevant perception-aware RL baseline---CPPO achieves notable gains. On the 3B model, CPPO improves average performance to 40.0\%, compared to PAPO's 38.1\%. On the larger 7B model, CPPO reaches 48.2\% versus PAPO's 46.8\%. Importantly, both CPPO and PAPO are trained under identical conditions---using the same dataset (ViRL39K) and the same number of training steps---ensuring that the improvements are not due to differences in data or compute. Thus, the consistent advantage of CPPO over PAPO can be attributed directly to the introduction of contrastive loss on perception tokens, which enhances the model's ability to capture and leverage visual information more effectively. The benefit of stronger visual grounding is what we expect to translate into improved reliability for downstream agentic tasks.

\begin{figure*}[t!]
    \centering
    \includegraphics[width=0.95\textwidth]{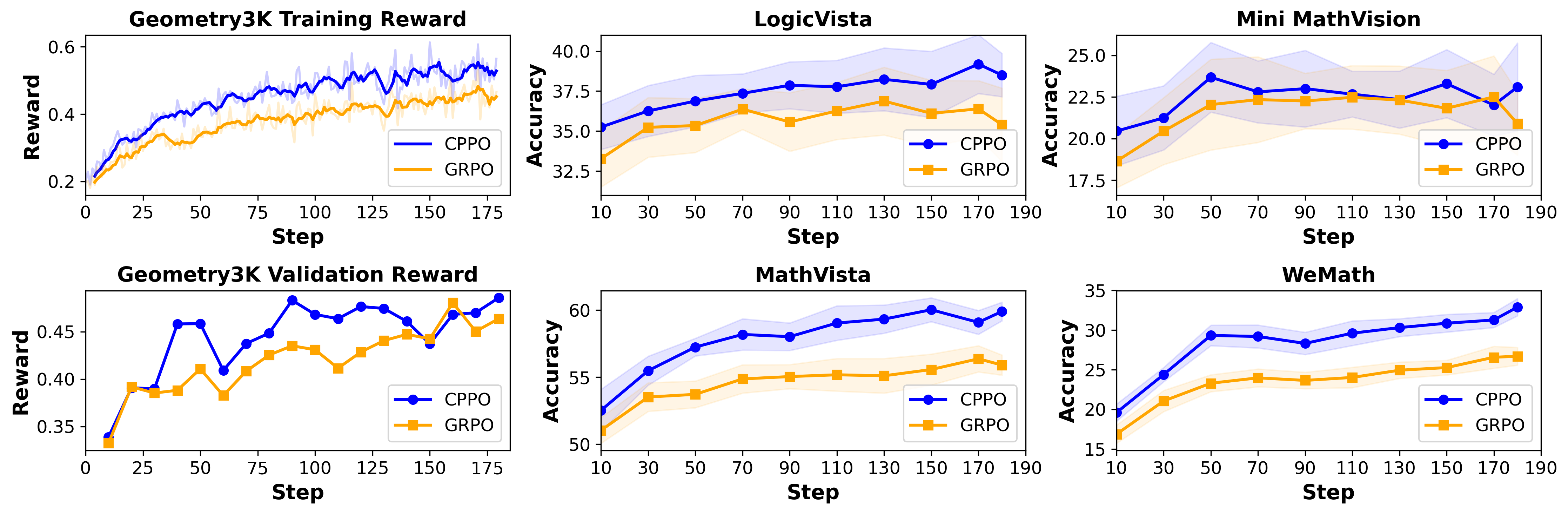}
    \caption{CPPO vs. GRPO (avg@8) on Qwen2.5-VL-3B across in-domain and out-of-domain scenarios. The X-axis represents RL training steps. The shaded area corresponds to one standard deviation over 8 responses. \textbf{1st column:} Reward comparison on the in-domain dataset during training. \textbf{2nd and 3rd columns:} Comparison on four out-of-domain visual reasoning benchmarks.}
    \label{fig:rewards_accuracy}
\end{figure*}

\textbf{Out-of-Domain Performance.}
Our main training is conducted on the ViRL39K \cite{vl-rethinker} dataset, a comprehensive dataset covering a wide range of topics.
To further evaluate the out-of-distribution performance of CPPO compared to GRPO, we additionally train Qwen2.5-VL-3B on the Geometry3K dataset \cite{geometry3k}, a small dataset with 2.1K samples focused on geometry problems.
We train the model using both GRPO and CPPO and evaluate the resulting models on out-of-distribution benchmarks.
\Cref{fig:rewards_accuracy} shows the training dynamics  (Training Reward), the reward on the in-domain validation set (Geometry3K Validation Reward), and the accuracy on out-of-distribution benchmarks as training progresses.
The training reward indicates that CPPO leads to faster learning and stronger generalization from the early stages of training.

\begin{figure*}[t!]
    \centering
    \includegraphics[width=0.92\linewidth]{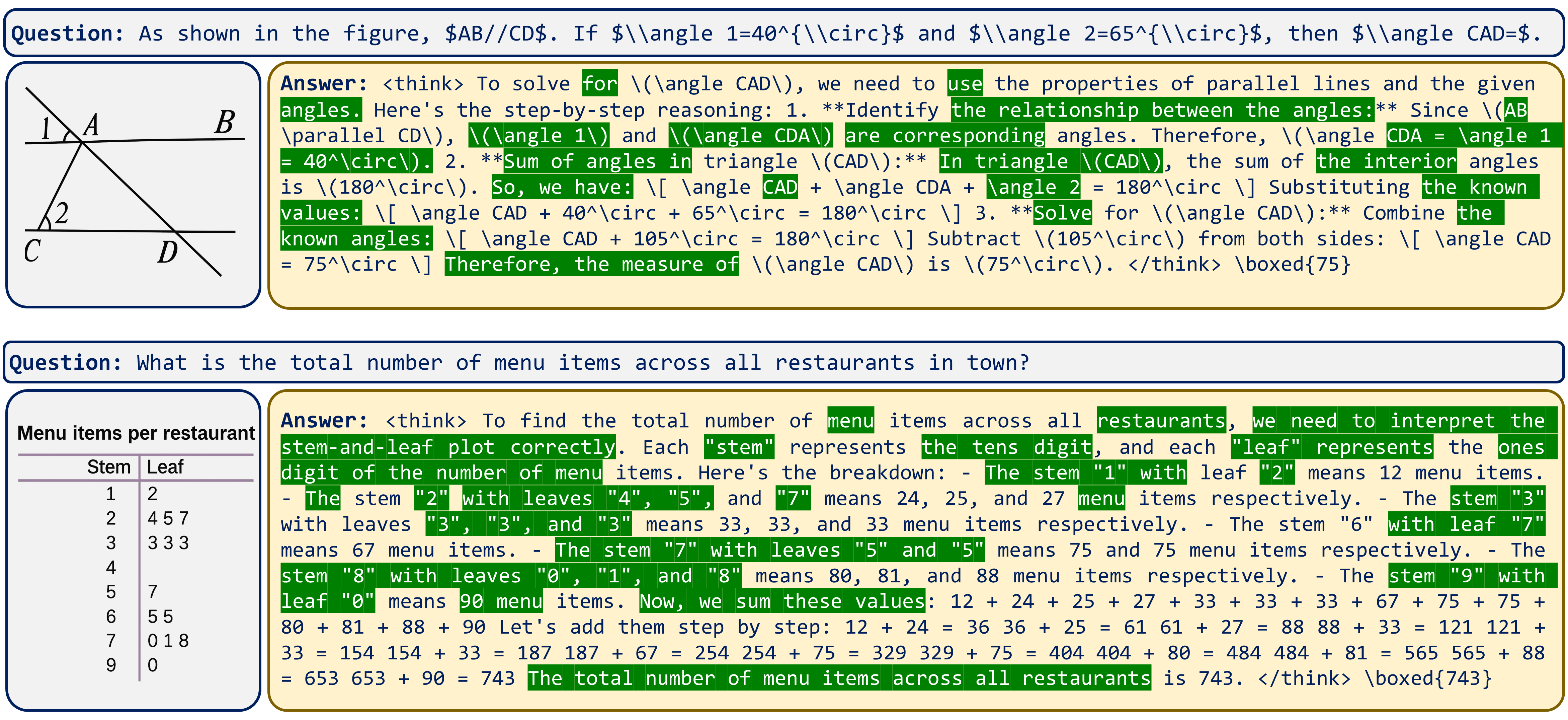}
    \caption{Sample outputs generated with CPPO with top 40\% detected perception tokens.}
    \label{fig:selected_tokens}
\end{figure*}

\textbf{Statistical Significance.}
\Cref{fig:rewards_accuracy} shows one-standard-deviation bands for all benchmarks, demonstrating that the gains of CPPO over GRPO are consistent across different evaluations.
To further validate this observation, we conduct paired t-tests for each benchmark.
All tests yield $\textit{p-values} \leq 0.02$, confirming that the improvements of CPPO over GRPO are statistically significant.

\textbf{Performance of Perception Token Detection.}
\Cref{fig:selected_tokens} shows two samples, the policy model's outputs generated by CPPO, and the top 40\% of perception tokens identified using our entropy-based  method. In the 1st example, the question asks for the value of angle $\angle CAD$ in a geometry problem. The key visual clues needed to solve this question are: (1) $\angle CDA = \angle 1 = 40^{\circ}$, (2) CAD forms a triangle, and (3) $\angle 2 = \angle ACD$. With these three pieces of information alone, one could solve the problem without referring back to the original figure. We observe that all these critical elements are successfully highlighted within the top 40\% of selected perception tokens. The 2nd example shows a stem and leaf plot summarizing the number of menu items per restaurant in a town, which is used to answer a question. Here, we find that most of the relevant numerical values are also captured within the top 40\% of detected perception tokens, illustrating that the method effectively identifies the essential visual information for the question. More analysis is given in Appendix \ref{appendix:perception}.

\subsection{Analysis of Selected Perturbations}

\textbf{Validating the Selected Perturbations.}
To assess whether the selected perturbations effectively preserve or remove information in images, we applied each perturbation to the images in four benchmarks. The Qwen2.5-VL-7B base model was then tested on every modified benchmark to measure changes in task performance attributable to these perturbations. As shown in \cref{tab:image_aug_effectiveness}, information-preserving augmentations reduced average model performance by less than 1.5\%, indicating that, for the majority of images, the critical information remains intact. In contrast, information-removing perturbations resulted in a substantial drop in average accuracy, exceeding 14\% across both removal strategies, indicating the elimination of salient information needed to answer the questions.

\begin{table}[t!]
\centering
\caption{Validating selected info preserving/removing perturbation. %Images of each benchmark are perturbed using each of the selected perturbation types. Performance remains close to the original (no-perturbation) baseline for information-preserving perturbations, but drops significantly for information-removing perturbations. Results are from Qwen2.5-VL-7B base model (avg@8).
}
\label{tab:image_aug_effectiveness}
\resizebox{\columnwidth}{!}{
\begin{tabular}{l|cccc|c}
    \toprule
    \rowcolor[gray]{0.94}
    \textbf{Image Perturbation Type}& \textbf{LogicVista} & \textbf{MVista\textsubscript{m}} & \textbf{MVision\textsubscript{m}} &  \textbf{WeMath} & \textbf{{AVG}} \\ \midrule
        \textit{Original Images} & 45.1 & 65.6 & 24.5 & 33.3 & 42.1 \\ \midrule
    \textbf{Information Preserving} \\
    \hspace{1mm} Color Jitter & 44.5 & 64.3 & 24.6 & 31.6 & 41.3 \\
    \hspace{1mm} Gaussian Blur & 44.2 & 64.4 & 25.3 & 31.6 & 41.4\\
    \hspace{1mm} Random Perspective &  44.4 & 64.0 & 25.3 & 31.1 & 41.2\\
    \hspace{1mm} Random Rotation & 43.9 & 63.4 & 24.0 & 31.4 & 40.7\\
    \midrule
    \textbf{Information Removing} \\
    \hspace{1mm} Random Occlusion & 31.3 & 40.9 & 19.3 & 13.8 & 26.3\\
    \hspace{1mm} Random Zoom Crop & 31.2 & 42.9 & 20.1 & 15.1 & 27.3\\
    \bottomrule
\end{tabular}}
\end{table}

\begin{table}[t!]
    \centering
    \caption{Impact of the perturbations used in CPPO. Co: Color, Ge: Geometry, Cr: Crop, Ma: Mask.}
    \label{tab:img_perturb_ablation}
    \resizebox{\columnwidth}{!}{
    \begin{tabular}{c|c|cccc|c}
        \toprule
        \rowcolor[gray]{0.94}
        \multicolumn{2}{c}{} & \textbf{LogicVista} & \textbf{MVista\textsubscript{m}} & \textbf{MVision\textsubscript{m}} & \textbf{WeMath} & \textbf{AVG} \\
        \midrule
        \multicolumn{2}{c}{\textbf{Perturb(+)} \,\,\,\,\,\,\,\,\:  \textbf{Perturb(-)}} &  &  &  &  \\
        Co & (Cr, Ma) & 37.7 & 56.9 & 22.4 & 31.1 & 37.0\\
        Ge & (Cr, Ma) & 37.8 & 58.7 & 22.0 & 32.5 & 37.8 \\
        (Ge, Co) & Ma & 38.7 & 58.8 & 22.6 & 32.3 & 38.1\\
        (Ge, Co) & Cr & 38.3 & 59.8 & 23.1 & 32.8 & 38.5\\
        (Ge, Co) & (Cr, Ma) & \textbf{38.5} & \textbf{59.9} & \textbf{23.1} & \textbf{32.9} & \textbf{38.6}\\
        \bottomrule
    \end{tabular}}
\end{table}

\begin{table}[h!]
    \centering
    \caption{Increase in training time and performance when training with CPPO for 2 epochs vs. GRPO for 4 epochs, relative to training with GRPO for 2 epochs. Even a 100\% increase in GRPO time does not match the performance achieved by CPPO.}
    \label{tab:additional_training_supp}
    \resizebox{\columnwidth}{!}{
    \begin{tabular}{l|cccc|l}
         \toprule
         %&\multicolumn{5}{c|}{\textbf{Math Benchmarks}}&\multicolumn{2}{c|}{\textbf{Visual Reasoning}}&\\ \cmidrule{2-9}
         \rowcolor[gray]{.9}
        %\textbf{Methods (Relative Increase of Training Time)} 
        & \textbf{MVista}\textsubscript{m}  & \textbf{WeMath} & \textbf{MVision\textsubscript{m}}  &\textbf{LogicVista}&\textbf{AVG} \\ \midrule
        GRPO: 2 Epochs               & 63.7 &  28.4 & 25.1 & 37.7 & 38.7\\
        GRPO: 4 Epochs (100\%)       & 65.8 &  28.5 & \textbf{25.4} & 38.2 & 39.5\\
        \textbf{CPPO: 2 Epochs (39\%)} & \textbf{66.3} &  \textbf{30.8} & \textbf{25.4} & \textbf{39.4} & \textbf{40.5}\\
        \bottomrule
    \end{tabular}}
\end{table}

\textbf{Impact of Selected Perturbations.}
We analyzed the effect of selected perturbations in the CPPO pipeline by training Qwen2.5-VL-3B on Geometry3K dataset \cite{geometry3k} with different selected perturbations. As shown in the \cref{tab:img_perturb_ablation}, a more diverse perturbation set provides a richer learning signal, hence leading to better performance.

% \begin{table}[h!]
%     \centering
%     \caption{Increase in training time and performance when training Qwen2.5-VL-3B with CPPO for 2 epochs or with GRPO for 4 epochs, relative to training with GRPO for 2 epochs. Even a 100\% increase in GRPO training time does not match the performance achieved by CPPO. All results are based on avg@8.}
%     \label{tab:additional_training_supp}
%     \resizebox{\columnwidth}{!}{
%     \begin{tabular}{l|ccccc|cc|l}
%          \toprule
%          &\multicolumn{5}{c|}{\textbf{Math Benchmarks}}&\multicolumn{2}{c|}{\textbf{Visual Reasoning}}&\\ \cmidrule{2-9}
%          \rowcolor[gray]{.9}
%         %\textbf{Methods (Relative Increase of Training Time)} 
%         & \textbf{MVista}\textsubscript{m} & \textbf{DMath} & \textbf{WeMath} & \textbf{MVision\textsubscript{m}} & \textbf{MVerse} & \textbf{MMMU-P\textsubscript{v}}&\textbf{LogicVista}&\textbf{AVG} \\ \midrule
%         GRPO: 2 Epochs               & 63.7 & 45.7 & 28.4 & 25.1 & 38.3 & 25.8 & 37.7 & 37.8\\
%         GRPO: 4 Epochs (100\%)       & 65.8 & 47.7 & 28.5 & \textbf{25.4} & 39.0 & 27.3 & 38.2 & 38.8\\
%         \textbf{CPPO: 2 Epochs (39\%)} & \textbf{66.3} & \textbf{48.9} & \textbf{30.8} & \textbf{25.4} & \textbf{39.4} & \textbf{28.5} & \textbf{40.9} & \textbf{40.0}\\
%         \bottomrule
%     \end{tabular}}
% \end{table}

\subsection{Complexity Analysis}
Compared to GRPO, CPPO adds additional computation due to two extra forward passes required to compute token distributions conditioned on the positive ($I^+$) and negative ($I^-$) images. This increases the time per training step. As shown in \cref{tab:additional_training_supp}, Training Qwen2.5-VL-3B with CPPO for two epochs takes 39\% more time than GRPO under identical resources.
One may argue that the same compute could instead be used to train GRPO longer. To test this, we trained Qwen2.5-VL-3B with GRPO for two additional epochs on ViRL39K \cite{vl-rethinker}. This doubles the training time (100\% increase) yet still underperforms CPPO trained for only two epochs (CPPO@2 epochs averages 40.5\% vs.\ GRPO@4 epochs at 39.5\%). These results indicate that CPPO's gains arise from the contrastive learning signal rather than increased compute.

\subsection{Ablations}
We adopt Qwen2.5-VL-3B as the baseline and conduct all ablations on the Geometry3K \cite{geometry3k}, which contains 2.1K samples. We select Geometry3K both to enable faster training and to demonstrate generalizability of CPPO.

\begin{table}[h!]
\centering
\caption{Ablation on $\lambda$ values.}
\label{tab:lambda_supp}
\resizebox{0.8\columnwidth}{!}{%
\begin{tabular}{l|cccc|c}
    \toprule
    \rowcolor[gray]{0.94}
    $\lambda$ & \textbf{LogicVista} & \textbf{MVista\textsubscript{m}} & \textbf{MVision\textsubscript{m}} &  \textbf{WeMath} & \textbf{AVG}\\
    \midrule
     0.01 & 37.4 & 59.2 & 21.9 & 31.4 & 37.5 \\
     0.02 & 38.5 & 59.9 & 23.1 & 32.9 & \textbf{38.6}\\
     0.03 & 38.6 & 57.8 & 22.9 & 28.8 & 37.0\\
     0.04 & 35.6 & 55.9 & 21.7 & 27.6 & 35.2\\
    \bottomrule
\end{tabular}}
\end{table}

\begin{table}[h!]
\centering
\caption{Ablation on top-$k$ perception tokens.}
\label{tab:topK_supp}
\resizebox{0.8\columnwidth}{!}{%
\begin{tabular}{l|cccc|c}
    \toprule
    \rowcolor[gray]{0.94}
    \textbf{K}& \textbf{LogicVista} & \textbf{MVista\textsubscript{m}} & \textbf{MVision\textsubscript{m}} &  \textbf{WeMath} & \textbf{AVG}\\
    \midrule
      5\% & 32.5 & 52.2 & 21.4 & 20.1 & 31.6 \\
     25\% & 36.7 & 57.9 & 22.7 & 30.7 & 37.0 \\
     50\% & 38.5 & 59.9 & 23.1 & 32.9 & \textbf{38.6}\\
     75\% & 37.6 & 57.4 & 22.3 & 29.1 & 36.6 \\
    100\% & 36.3 & 56.9 & 22.0 & 29.5 & 36.2 \\
    \bottomrule
\end{tabular}}
\end{table}

\textbf{Loss Weighting ($\lambda$).}
We experiment with different $\lambda$ values in \cref{eq:objective_func}. $\lambda$ controls the strength of perceptual grounding. As given in \cref{tab:lambda_supp},  the best performance is obtained with $\lambda=0.02$, achieving 38.6\% average accuracy. In general, CPPO with different $\lambda$ values outperforms GRPO with an average accuracy of 34.7\%.

\textbf{Top$k$.}
\Cref{tab:topK_supp} presents the analysis of different $K$ values for top$k$ perception token detection. The results show that average accuracy improves as $K$ increases from 5\% to 50\%, but declines when $K$ is further expanded from 50\% to 100\%.
We hypothesize that this trend arises because larger $K$ values include more tokens that the policy model is already confident about (i.e., tokens with lower entropy change), which are less informative perception tokens. Incorporating these tokens can slow down the training and ultimately lead to worse performance when models are trained for the same number of epochs. %Detailed per-benchmark numbers are in Table~2 of the supplementary materials. 
More ablations on different components of CPPO are given in Appendix \ref{appendix:ablation_components}.

%% file: 5_conclusion.tex
\section{Conclusion}
In this work, we introduced CPPO, a perception-aware RL-based method for finetuning VLMs. CPPO leverages an entropy-based approach to disentangle \textit{perception} tokens from \textit{reasoning} tokens, where perception tokens capture visual information extracted from the input image. To better align training with perception quality, we proposed a Contrastive Perception Loss (CPL)---an unsupervised, model-free objective that penalizes perception errors. Extensive experiments demonstrate that CPPO outperforms recent RL methods for VLMs, achieving state-of-the-art performance across multiple math and visual reasoning benchmarks. Because perception failures are a primary driver of unsafe behavior in vision--language agents, we believe a self-supervised perception-aware RL recipe like CPPO is a promising building block for training agents that must operate reliably in open-ended environments. Limitations and additional analysis are discussed in the Appendix.

%% file: X_appendix.tex
\section{Proof for Proposition 1}
\label{appendix:proof-vision-entropy}
\textbf{Proposition 1 (Entropy increase as a proxy for vision dependence).} \textit{Let $I$ denote the original image, $I^{-}$ a perturbed variant that removes query-relevant perceptual information, and $\mathbf{o}_i$ the sequence of tokens generated by the policy when conditioned on $I$. The increase in entropy of a token $o_{i,t} \in \mathbf{o}_i$, when the policy is conditioned on $I^{-}$ rather than $I$, serves as a proxy for the degree to which the policy model associates $o_{i,t}$ with the query-relevant visual content of $I$. This increase is calculated as follows:}
\begin{equation*}
    \Delta H_{i,t} = H(o_{i,t}|q,I^{-},\mathbf{o}_{i,<t}) - H(o_{i,t}|q,I,\mathbf{o}_{i,<t}).
\end{equation*}
\begin{proof}
Recall the identity relating conditional mutual information (denoted by $MI$) and conditional entropy:
\begin{equation}
\begin{split}
    &H\!\left(o_{i,t} \mid X, q, \mathbf{o}_{i,<t}\right) \;=\;\\
    &H\!\left(o_{i,t} \mid q, \mathbf{o}_{i,<t}\right) \;-\;
    MI\!\left(o_{i,t} ; X \,\middle|\, q, \mathbf{o}_{i,<t})\right).
\end{split}
\end{equation}
Applying this with both $X=I$ and $X=I^{-}$ and subtracting, we obtain
\begin{equation}
\label{eq:delta-entropy-as-mi-gap}
\begin{split}
&H_{i,t}(I^{-}) - H_{i,t}(I)
\;=\; \\
&H\!\left(o_{i,t} \mid I^{-}, q, \mathbf{o}_{i,<t}\right) - H\!\left(o_{i,t} \mid I, q, \mathbf{o}_{i,<t}\right)
\;=\; \\
&MI\!\left(o_{i,t} ; I \,\middle|\, q, \mathbf{o}_{i,<t}\right)
\;-\;
MI\!\left(o_{i,t} ; I^{-} \,\middle|\, q, \mathbf{o}_{i,<t}\right).
\end{split}
\end{equation}
$I^{-}$ is obtained from $I$ by an information-removing augmentation that obscures query-relevant visual information. Our main assumption is that the conditional mutual information between perception tokens in $\mathbf{o}_i$ and $I$ should be greater than their conditional mutual information with the perturbed image $I^{-}$. Formally, if $o_{i,t}$ is a perception token, we assume the following inequality holds for its conditional mutual information:
\begin{equation}
MI\!\left(o_{i,t} ; I \,\middle|\, q, \mathbf{o}_{i,<t}\right) - MI\!\left(o_{i,t} ; I^{-} \,\middle|\, q, \mathbf{o}_{i,<t}\right)
\;\ge\; 0.
\end{equation}
Substituting this inequality into \eqref{eq:delta-entropy-as-mi-gap} yields
\begin{equation}
H_{i,t}(I^{-}) - H_{i,t}(I) \;\ge\; 0.
\end{equation}
Thus, an increase in predictive entropy, $\Delta H_{i,t}$, serves as a principled proxy for identifying vision-dependent tokens in the output sequence.
\end{proof}

\section{Analysis on Performance of Perception Token Detection}
\label{appendix:perception}
\begin{figure}[h!]
    \centering
    \includegraphics[width=0.85\columnwidth]{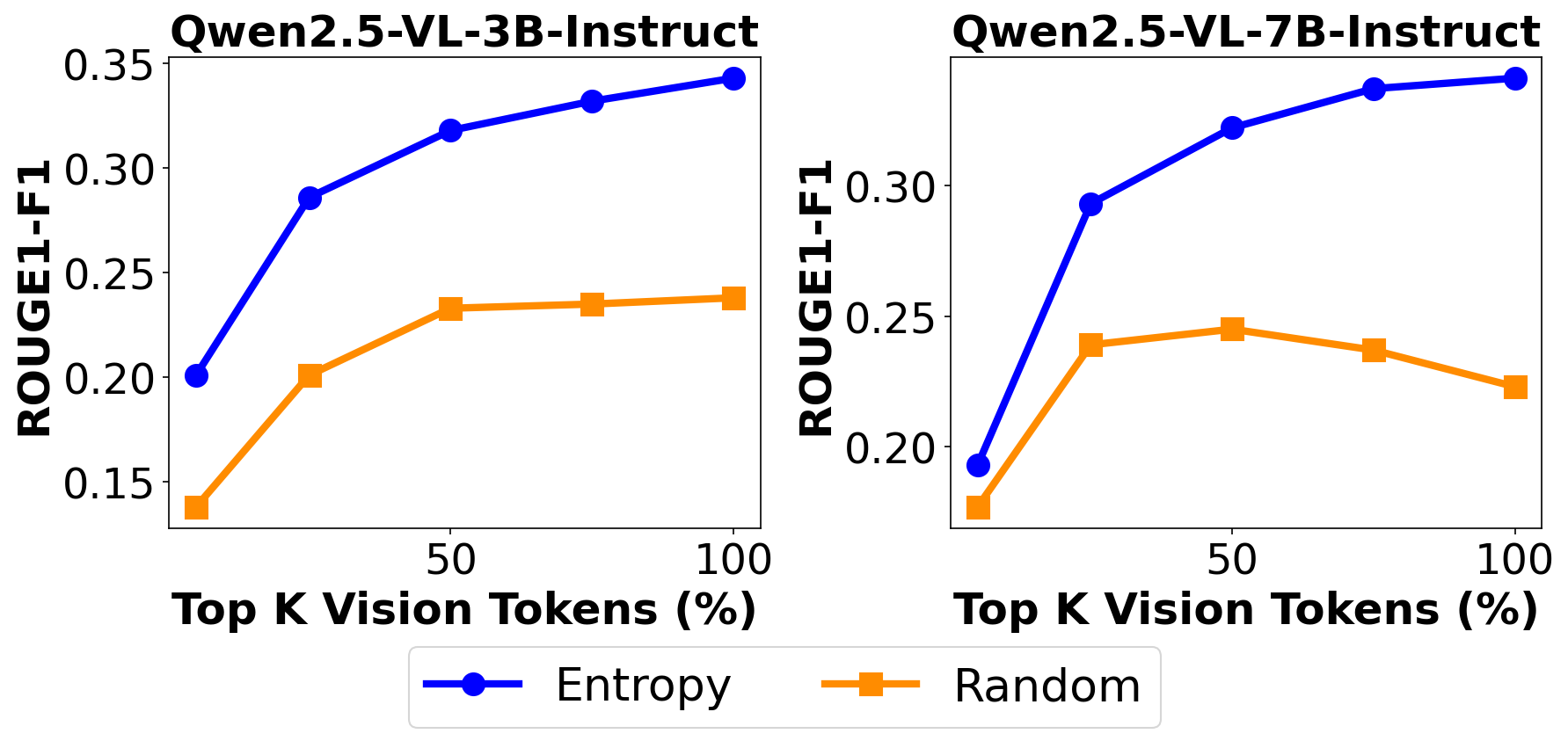}
    \caption{Quantitative evaluation of perception token detection.}
    \label{fig:vision_detection_accuracy}
\end{figure}

To quantitatively evaluate our perception detection method, we used the inference outputs of Qwen2.5-VL-3B and -7B
on four test sets: MathVista-MINI, LogicVista, MathVision-MINI, and WeMath. We then passed these outputs to GPT5-mini, which was used to separate the perception-related information from the rest of the model's response. This extracted perception information serves as our ground truth. We measure the accuracy of our detection method by calculating the ROUGE-1 F1 score between the detected perception tokens and the GPT5-mini outputs.
It is important to note that GPT5-mini's separation is not flawless; thus, this evaluation should be viewed as a proof-of-concept rather than a definitive benchmark.
Figure \ref{fig:vision_detection_accuracy} shows that the ROUGE-1 F1 score improves as we increase the number of top$k$ perception tokens, up to the point where 100\% of perception tokens are included. Here, 100\% refers to selecting all tokens with positive $\Delta H$ in Proposition 1, rather than all output tokens. At each top$k$ percentage, we also select the same number of tokens randomly to serve as a baseline. Figure \ref{fig:vision_detection_accuracy} shows that there is significant gap between our entropy-based method and random selection.

\section{Extra Training Details}
\label{appendix:details}
Table \ref{tab:hyper} shows the summary of hyper-parameters used in training of 3B and 7B models.

\begin{table}[h!]
\centering
\caption{Summary of training hyperparameter configurations.}
\label{tab:hyper}
\small
\begin{tabular}{ll}
\toprule
\textbf{Parameter} & \textbf{Configuration} \\
\midrule
\multicolumn{2}{l}{\textbf{Main Results}} \\
Model Base & Qwen2.5-VL-Instruct \\
Global Batch Size & 512 \\
Rollout Temperature & 1.0 \\
Learning Rate & $1e^{-6}$ \\
Rollout Number & 5 \\
Training Epochs & 2 \\
Optimizer & AdamW \\
Policy Loss Aggregation & \texttt{token-mean} \\
$\beta$ & 0.01\\
$\tau$ & 0.1\\
$k$ & 50\% \\
$\lambda$ & 0.02\\
\midrule
\multicolumn{2}{l}{\textbf{Ablations Specific}} \\
Dataset & Geometry3K\\
Training Epochs & 12\\
Global Batch Size & 128 \\
\bottomrule
\end{tabular}
\end{table}

\section{Ablation on Main Components of CPPO}
\label{appendix:ablation_components}
\Cref{tab:ablation} reports the ablation study on the key components of CPPO. Starting from GRPO, applying CPL to all tokens raises the average accuracy from 34.7\% to 35.0\%. Restricting CPL to only the top 50\% of perception tokens yields a larger gain, increasing accuracy to 36.6\%. Finally, introducing advantage gating---where the contrastive loss is applied only to rollouts with positive advantage---further improves performance to 38.6\%.
These results highlight that each component makes a meaningful contribution, and together they account for the overall effectiveness of CPPO.

\begin{table}[h!]
\centering
\caption{Ablation on the components of CPPO.}
\label{tab:ablation}
\resizebox{\columnwidth}{!}{%
\begin{tabular}{l|cccc|c}
    \toprule
    \rowcolor[gray]{0.94}
    \textbf{Methods}& \textbf{LogicVista} & \textbf{MVista\textsubscript{m}} & \textbf{MVision\textsubscript{m}} &  \textbf{WeMath} & \textbf{AVG} \\ \midrule
    \textit{Qwen2.5-VL-3B} & 32.4 & 56.4 &19.5 & 14.5 & 30.7\\ \midrule
    GRPO & 35.4 & 55.9& 20.9& 26.7 & 34.7 \\
     \hspace{1mm} + CPL on All Tokens & 35.6 & 56.0 & 20.8 & 27.2 & 35.0 \\
     \hspace{1mm} + CPL on Top$k$ Perc. Tokens & 36.4 & 56.6 & 22.5 & 30.9 & 36.6 \\
     \hspace{1mm} + Advantage Gating & 38.5 & 59.9 & 23.1 & 32.9 & \textbf{38.6} \\
    \bottomrule
\end{tabular}}
\end{table}

\section{Image Perturbation Details}
\label{appendix:perturbations}
\begin{figure}[h!]
  \centering
  \includegraphics[width=0.98\linewidth]{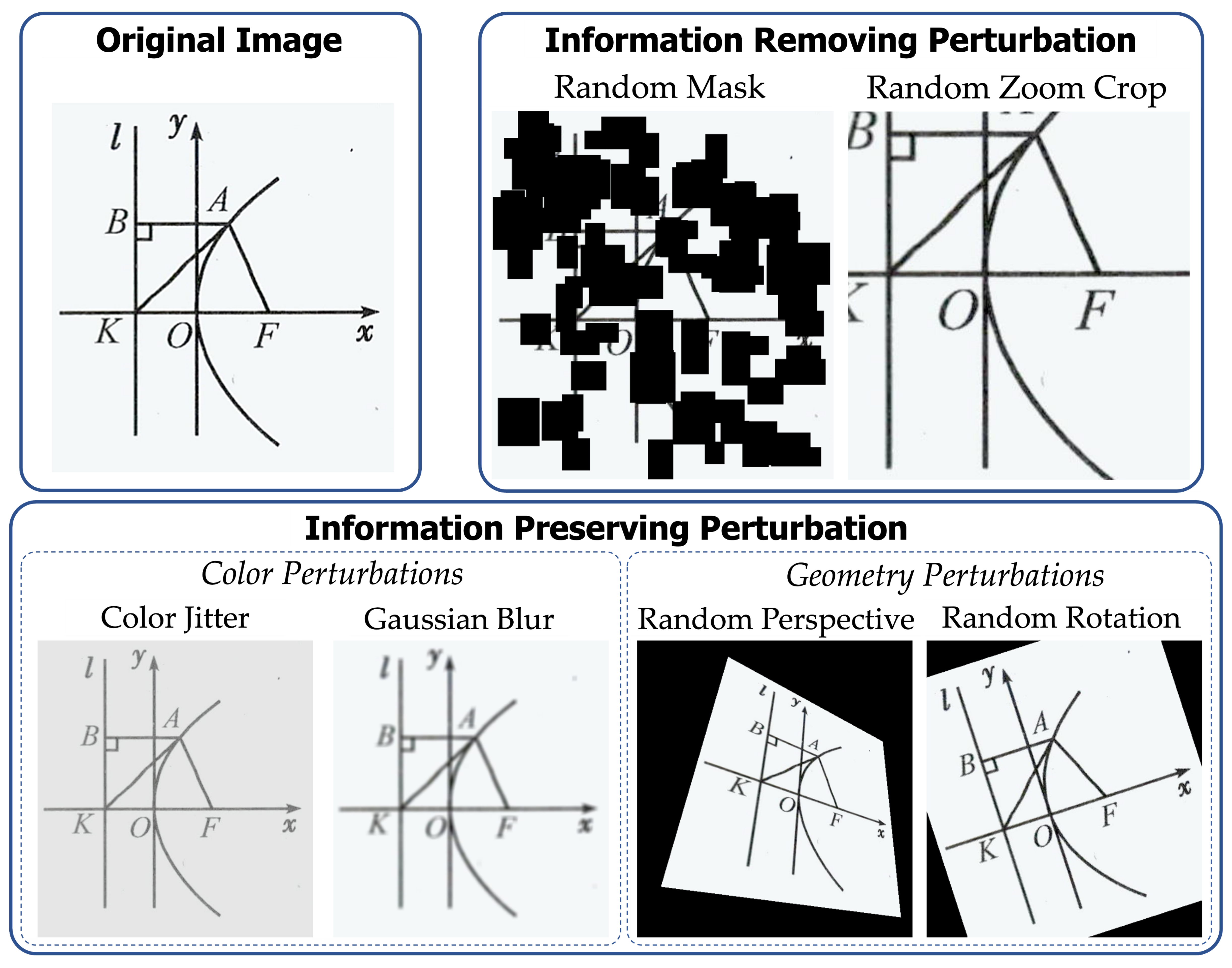}
  \caption{Samples of information-removing and information-preserving perturbations used in CPPO pipeline.}
  \label{fig:perturbation}
\end{figure}

The information-removing perturbations, such as random occlusion and random zoom crop, eliminate key visual details necessary for understanding the image. In contrast, the information-preserving perturbations---including color jitter, random perspective, random rotation, and Gaussian blur---modify the image without discarding critical information. Table \ref{tab:aug_parameters} shows the Torchvision parameters selected for each perturbation. Samples of information-removing and information-preserving perturbations used in CPPO pipeline are also shown in \cref{fig:perturbation}.

\begin{table}[h!]
\centering
\caption{Selected image perturbation parameters.}
\label{tab:aug_parameters}
\small
\begin{tabular}{ll}
\toprule
\textbf{Perturbation} & \textbf{Parameters} \\
\midrule
Color Jitter &  Brightness: $(0.2, 1.3)$ \\
& Contrast: $(0.2, 1.8)$ \\
& Saturation: $(0.2, 1.8)$ \\
Random Perspective & Distortion Scale: $0.2$  \\
Random Rotation & Degrees: $10$ \\
Gaussian Blur & Kernel Size: $3$ \\ \midrule
Random Occlusion & 80\% Patch-wise Masking\\
Random Zoom Crop & Retain 30\% of Image\\
\bottomrule
\end{tabular}
\end{table}

\section{Qualitative Results}
\label{appendix:qualitative}
\begin{figure*}[h!]
    \centering
\includegraphics[width=0.8\textwidth]{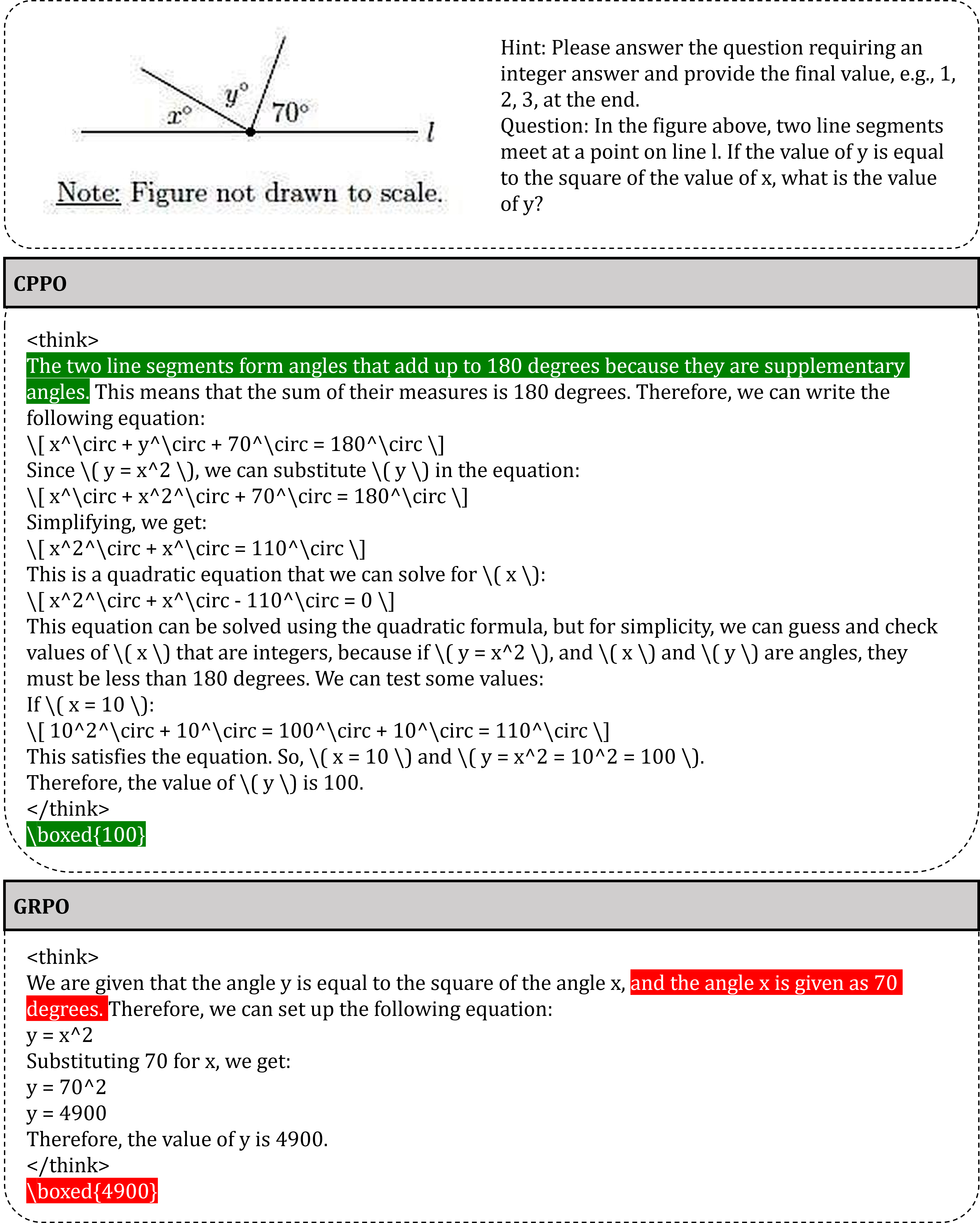}
    \caption{Sample generated responses by CPPO and GRPO. GRPO exhibits a perception error that is corrected in the CPPO response.}
    \label{fig:appendix_qualit_1}
\end{figure*}

\begin{figure*}[h!]
    \centering
\includegraphics[width=0.85\textwidth]{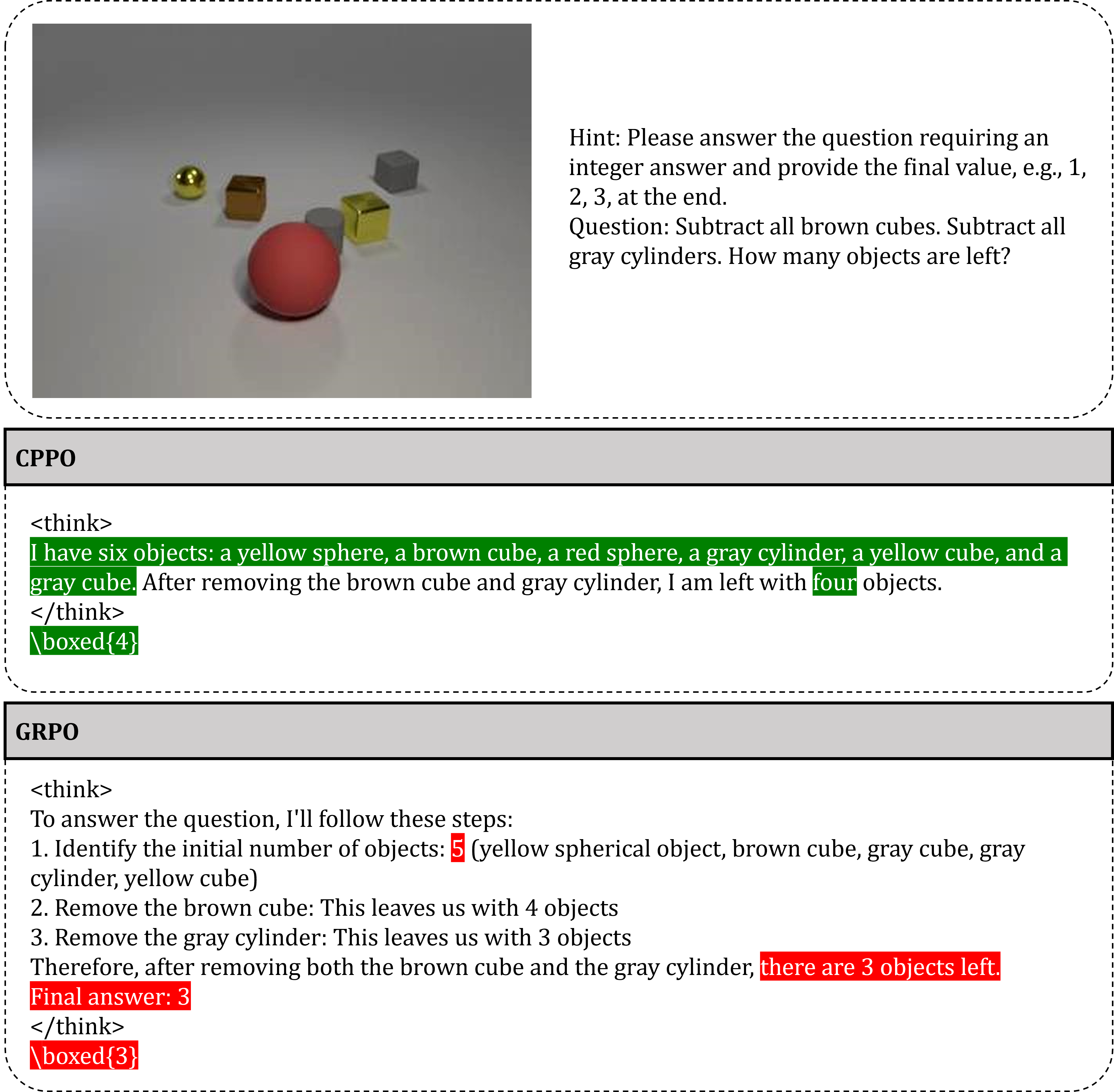}
    \caption{Sample generated responses by CPPO and GRPO. GRPO exhibits a perception error that is corrected in the CPPO response.}
    \label{fig:appendix_qualit_2}
\end{figure*}

\begin{figure*}[h!]
    \centering
\includegraphics[width=0.85\textwidth]{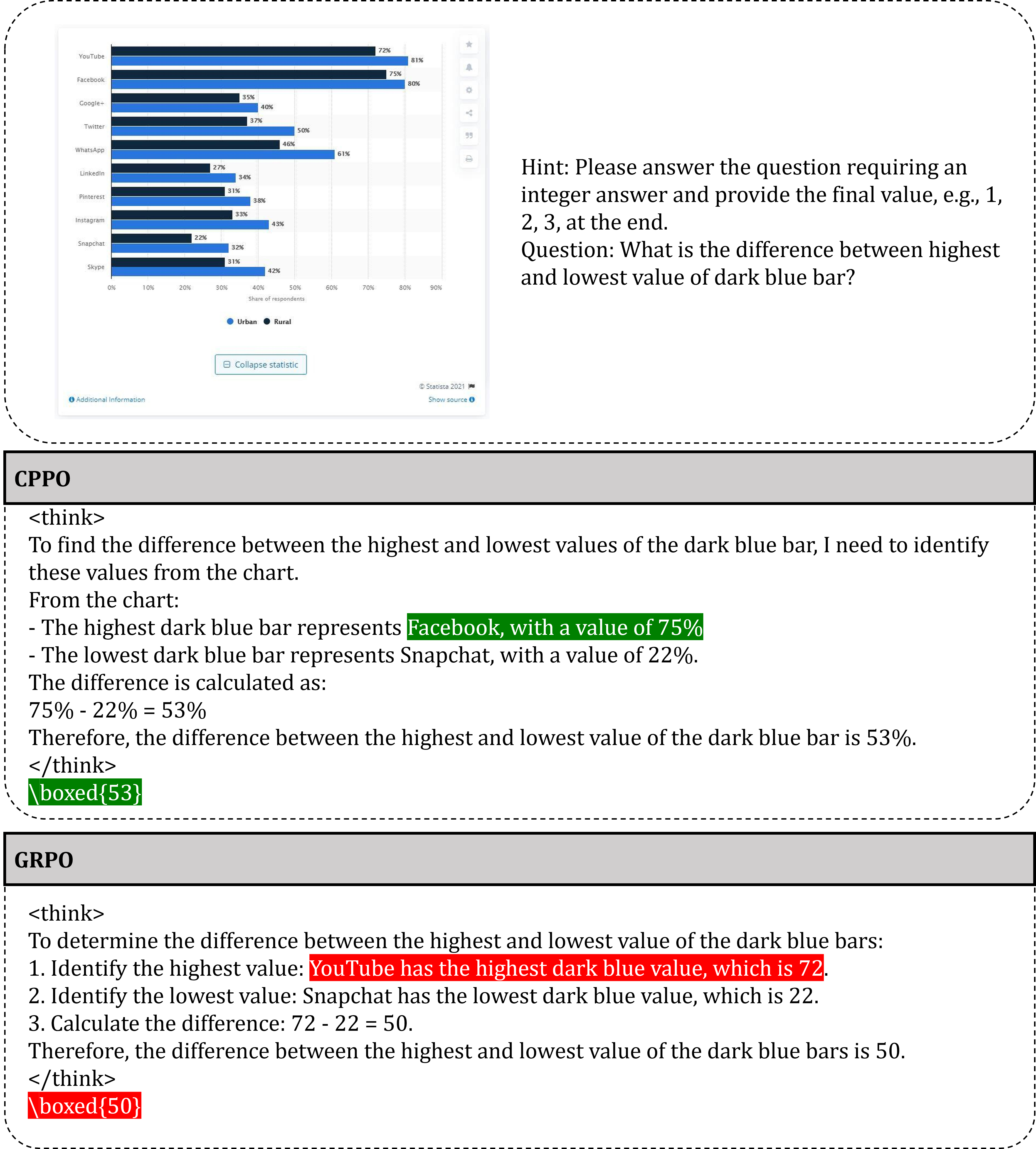}
    \caption{Sample generated responses by CPPO and GRPO. GRPO exhibits a perception error that is corrected in the CPPO response.}
    \label{fig:appendix_qualit_3}
\end{figure*}

Figures \ref{fig:appendix_qualit_1}--\ref{fig:appendix_qualit_3} show three qualitative examples. We observe that CPPO has corrected the perception mistakes of models trained with GRPO. For example, in Figure \ref{fig:appendix_qualit_1}, the model trained with GRPO states that ``the angle x is given as 70 degrees'' that is a wrong perception information extracted from the image. However, the model trained with CPPO corrected this statement by ``The two line segments form angles that add up to 180 degrees''. Note that when perception tokens are wrong, even with correct reasoning trajectory, the final answer is wrong.

\section{Limitations}
\label{appendix:limitations}
This work has several limitations that should be addressed in future research. First, due to our computational constraints, we did experiments up to 3B and 7B models. Exploring larger VLMs, such as 72B models, is an important direction for future work. Second, our evaluation was limited to Qwen2.5-VL baselines; extending the analysis to other baselines, such as InternVL \cite{chen2024internvl}, would provide a more comprehensive comparison. While we demonstrated the effectiveness of CPPO using 40K training samples, future studies should investigate large-scale training with substantially larger datasets. Finally, our experiments target standard multimodal reasoning benchmarks; we did not yet evaluate CPPO-trained policies inside actual agent harnesses (web/computer-use environments, robotic-style perception loops). We hypothesize that the improvements in visual grounding will translate into reduced hallucinated actions in such settings, but leave that empirical evaluation to future work.